\documentclass[twocolumn, 9pt, DIV24]{scrartcl}
\pdfoutput=1
\usepackage[T1]{fontenc}
\addtokomafont{pagenumber}{\small\sffamily}

\usepackage{balance}
\usepackage{hyperref}
\usepackage{graphicx,dblfloatfix}
\usepackage{textcomp}
\usepackage{ctable}
\usepackage{floatrow}

\usepackage{url}

\usepackage[noend]{algpseudocode}
\usepackage{algorithm}
\algnewcommand{\LineComment}[1]{\State \(\triangleright\) \emph{#1}}

\usepackage[font={rm},labelfont={rm,bf},small,]{caption}
\setcapindent{0cm}
\DeclareCaptionLabelSeparator{pp}{| }

\DeclareCaptionType{SI-figure}[Fig.]
\DeclareCaptionLabelFormat{bf-parens-si}{\bfseries Extended Data Figure #2 }
\DeclareCaptionLabelFormat{bf-parens-table}{\bfseries Extended Data Table #2}
\usepackage{endnotes}
\usepackage{enumitem}
\usepackage{palatino}
\usepackage{amsmath,amsfonts,amssymb}
\usepackage{multirow}
\usepackage{sectsty}
\usepackage{setspace}
\usepackage{xr}
\usepackage{soul}
\usepackage{enumitem}

\usepackage[defernumbers=true,
  maxbibnames=1000,
  maxnames=30,
  backend=bibtex,
  firstinits=true,
  sorting=none,citetracker,
  url=false]{biblatex}
\renewcommand{\cite}{\supercite}
\defbibheading{bibliography}{}
\DeclareFieldFormat{labelnumberwidth}{\mkbibbold{#1.}}

\makeatletter
\patchcmd{\blx@citation@entry}
  {\ifinlistcs{#1}{blx@segm@\the\c@refsection @\the\c@refsegment}
     {}
     {\listcsgadd{blx@segm@\the\c@refsection @\the\c@refsegment}{#1}}}
  {\ifboolexpr{ test {\blx@ifentryseen@global{#1}}
     or test {\ifinlistcs{#1}{blx@segm@\the\c@refsection @\the\c@refsegment}} }
     {}
     {\listcsgadd{blx@segm@\the\c@refsection @\the\c@refsegment}{#1}}}
  {}{}
\makeatother

\setlist{noitemsep}
\setlist{nolistsep}

\newcommand\blfootnote[1]{%
  \begingroup
  \renewcommand\thefootnote{}\footnote{#1}%
  \addtocounter{footnote}{-1}%
  \endgroup
}

\RequirePackage{fancyhdr}
\title{Robots that can adapt like animals} 

\author{Antoine Cully, Jeff Clune, Danesh Tarapore, Jean-Baptiste Mouret}

\date{}

\begin{document}
\begin{refsegment}
\pagestyle{fancy}
\twocolumn[{
\textbf{\textsf{This manuscript is the pre-submission manuscript provided by the authors.\\ For the final, post-review version, please see: \url{http://dx.doi.org/10.1038/nature14422}}\\
}

\hrule width \hsize \kern 0.5mm 
\hrule width \hsize \kern 0.5mm 
\hrule width \hsize height 0.5mm 

\begin{@twocolumnfalse}

\begin{flushleft}
    \fontsize{23}{8}\selectfont
  \bfseries Robots that can adapt like animals
  \end{flushleft}
  	Antoine Cully,$^{1,2}$ Jeff Clune,$^6$ Danesh Tarapore,$^{1,2}$ Jean-Baptiste Mouret$^{1-5,*}$

  	\bigskip

  \end{@twocolumnfalse}
  }]

\blfootnote{\sffamily $^1$ Sorbonne Universit\'es, UPMC Univ Paris 06, UMR 7222, ISIR, F-75005, Paris}
\blfootnote{\sffamily $^2$ CNRS, UMR 7222, ISIR, F-75005, Paris, France}
\blfootnote{\sffamily $^3$ Inria, Villers-l\`es-Nancy, F-54600, France}
\blfootnote{\sffamily $^4$ CNRS, Loria, UMR 7503, Vand\oe{}uvre-l\`es-Nancy, F-54500, France}
\blfootnote{\sffamily $^5$ Universit\'e de Lorraine, Loria, UMR 7503, Vand\oe{}uvre-l\`es-Nancy, F-54500, France}
\blfootnote{\sffamily $^6$ University of Wyoming, Laramie, WY, USA}
\blfootnote{\sffamily $^*$ Corresponding author: \href{mailto:jean-baptiste.mouret@inria.fr}{jean-baptiste.mouret@inria.fr}}

\vspace{-1.2em}
\bfseries
\noindent As robots leave the controlled environments of factories to autonomously function in more complex, natural environments\cite{Bellingham2007,yoerger2008underwater,broadbent2009acceptance}, they will have to respond to the inevitable fact that they will become damaged\cite{sanderson2010mars,carlson2005ugvs}. However, while animals can quickly adapt to a wide variety of injuries, current robots cannot ``think outside the box'' to find a compensatory behavior when damaged: they are limited to their pre-specified self-sensing abilities, can diagnose only anticipated failure modes\cite{blanke2006diagnosis}, and require a pre-programmed contingency plan for every type of potential damage, an impracticality for complex robots\cite{sanderson2010mars,carlson2005ugvs}. Here we introduce an intelligent trial and error algorithm that allows robots to adapt to damage in less than two minutes, without requiring self-diagnosis or pre-specified contingency plans. Before deployment, a robot exploits a novel algorithm to create a detailed map of the space of high-performing behaviors: This map represents the robot's intuitions about what behaviors it can perform and their value. If the robot is damaged, it uses these intuitions to guide a trial-and-error learning algorithm that conducts intelligent experiments to rapidly discover a compensatory behavior that works in spite of the damage. Experiments reveal successful adaptations for a legged robot injured in five different ways, including damaged, broken, and missing legs, and for a robotic arm with joints broken in 14 different ways. This new technique will enable more robust, effective, autonomous robots, and suggests principles that animals may use to adapt to injury.

\normalfont

\smallskip

\begin{figure*}
\begin{center}
\includegraphics[width=0.8\textwidth]{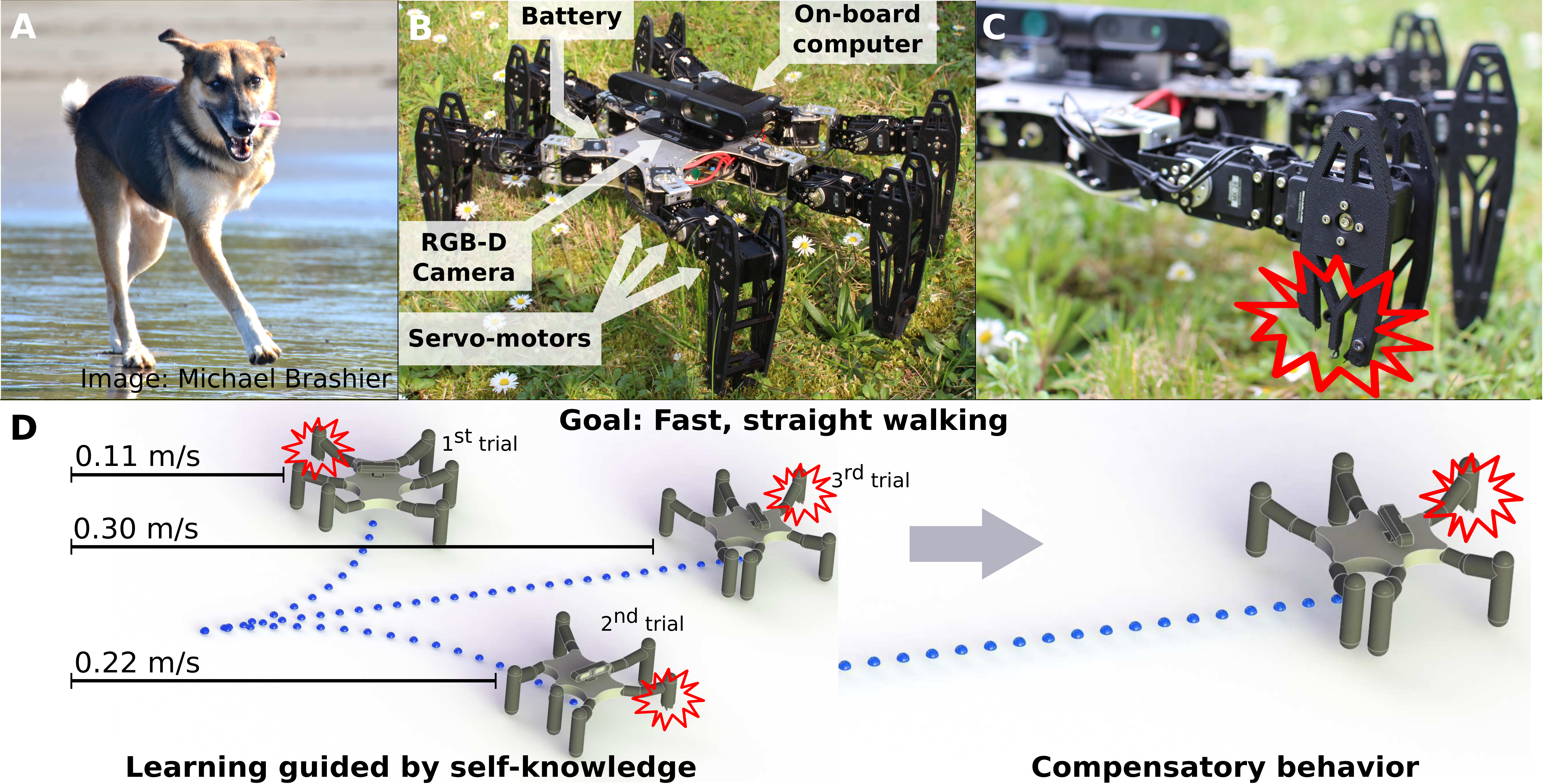}
\end{center}
\caption{\label{fig:concept}\textbf{With Intelligent Trial and Error, robots, like animals, can quickly adapt to recover from damage.} \textbf{(A)} Most animals can find a compensatory behavior after an injury. Without relying on \emph{predefined} compensatory behaviors, they learn how to avoid behaviors that are painful or no longer effective. \textbf{(B)} An undamaged, hexapod robot. \textbf{(C)} One type of damage the hexapod may have to cope with (broken leg). \textbf{(D)} After damage occurs, in this case making the robot unable to walk straight, damage recovery via Intelligent Trial and Error begins. The robot tests different types of behaviors from an automatically generated map of the behavior-performance space. After each test, the robot updates its predictions of which behaviors will perform well despite the damage. This way, the robot rapidly discovers an effective compensatory behavior.} 
\end{figure*}

Robots have transformed the economics of many industries, most notably manufacturing\nolinebreak[4]\cite{Siciliano2008}, and have the power to deliver tremendous benefits to society, such as in search and rescue\cite{murphy2004trial}, disaster response\cite{nagatani2013emergency}, health care\cite{broadbent2009acceptance}, and transportation\cite{thrun2006stanley}. They are also invaluable tools for scientific exploration, whether of distant planets\cite{Bellingham2007,sanderson2010mars} or deep oceans\cite{yoerger2008underwater}. A major obstacle to their widespread adoption in more complex environments outside of factories is their fragility\cite{sanderson2010mars,carlson2005ugvs}: Robots presently pale in comparison to natural animals in their ability to invent compensatory behaviors after an injury~(Fig.~\ref{fig:concept}A).

Current damage recovery in robots typically involves two phases: self-diagnosis, and then selecting the best, pre-designed contingency plan\cite{verma2004real,bongard2006resilient,fenton2001fault,Apollo13}. Such self-diagnosing robots are expensive, because self-monitoring sensors are expensive, and are difficult to design, because robot engineers cannot foresee every possible situation: this approach often fails either because the diagnosis is incorrect\cite{bongard2006resilient,fenton2001fault} or because an appropriate contingency plan is not provided\cite{Apollo13}.

Injured animals respond differently: they learn by trial and error how to compensate for damage (e.g. learning which limp minimizes pain)\cite{Jarvis2013,Fuchs2014}. Similarly, trial-and-error learning algorithms could allow robots to creatively discover compensatory behaviors without being limited to their designers' assumptions about how damage may occur and how to compensate for each damage type. However, state-of-the-art learning algorithms are impractical because of the ``curse of dimensionality''\cite{Kober2013}: the fastest algorithms constrain the search to a few behaviors (e.g. tuning only 2 parameters, requiring 5-10 minutes) or require human demonstrations\cite{Kober2013}. Algorithms without these limitations take several hours\cite{Kober2013}. Damage recovery would be much more practical and effective if robots adapted as creatively and quickly as animals (e.g. in minutes) and without expensive self-diagnosing sensors.


Here, we show that rapid adaptation can be achieved by guiding an intelligent trial-and-error learning algorithm with an automatically generated, pre-computed, behavior-performance map that predicts the performance of thousands of different behaviors (Supplementary Video S1). The key insight is that, whereas current learning algorithms either start with no knowledge of the search space\cite{Kober2013} or with minimal knowledge from a few human demonstrations\cite{Kober2013,argall2009survey}, animals better understand the space of possible behaviors and their value from previous experience\cite{Thelen1995}, enabling  injured animals to intelligently select tests that validate or invalidate whole families of promising compensatory behaviors.

We have robots store knowledge from previous experience in the form of a map of the behavior-performance space. Guided by this map, a damaged robot tries different types of behaviors that are predicted to perform well and, as tests are conducted, updates its estimates of the performance of those types of behaviors. The process ends when the robot predicts that the most effective behavior has already been discovered. The result is a robot that quickly discovers a way to compensate for damage (e.g. ~Fig.~\ref{fig:concept}C) without a detailed mechanistic understanding of its cause, as occurs with animals. We call this approach ``Intelligent Trial and Error'' (Fig.~\ref{fig:concept}D).

\begin{figure*}
\begin{center}
\includegraphics[width=0.7\textwidth]{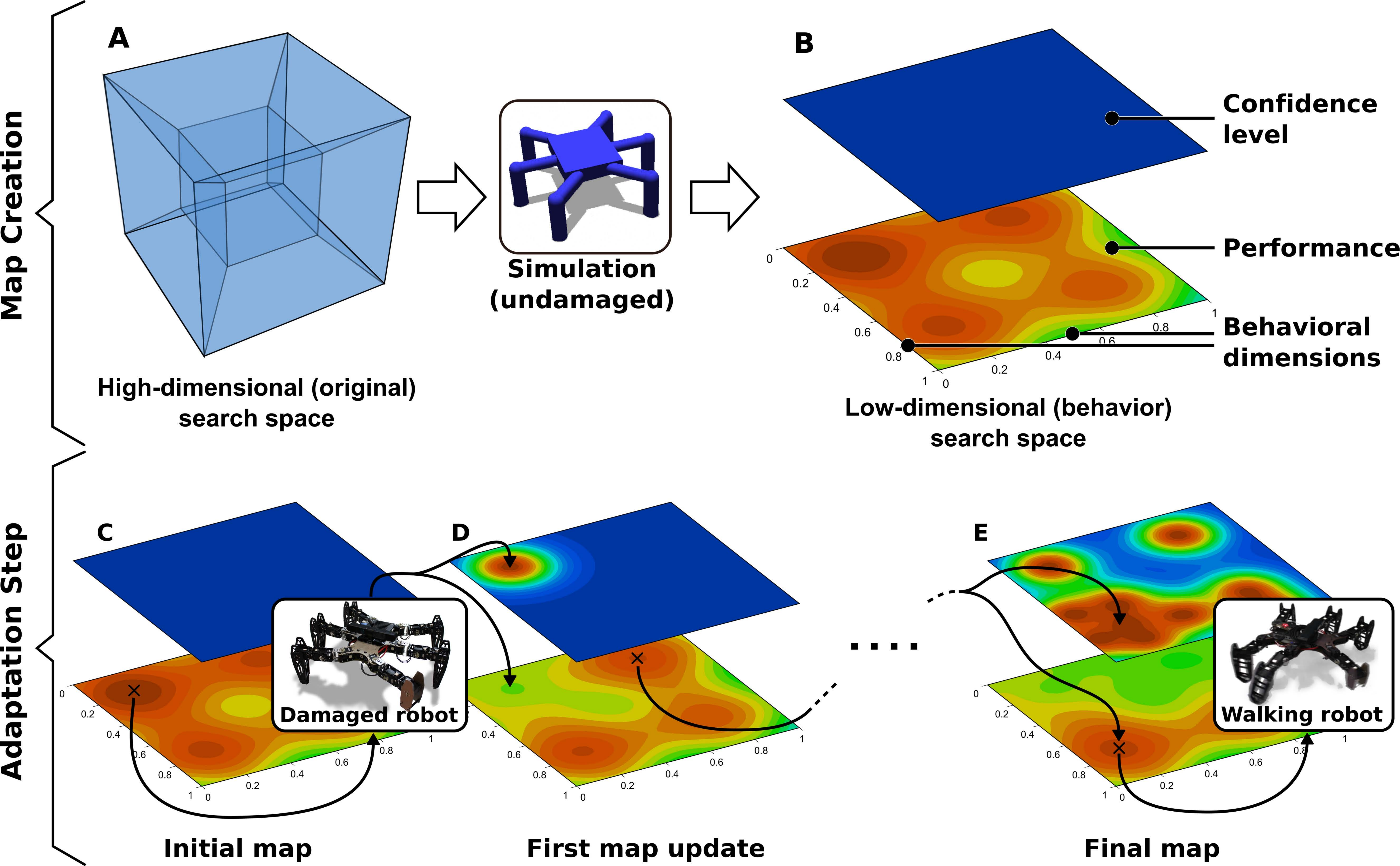}
\end{center}
\caption{\label{fig:bo}\textbf{(A \& B). Creating the behavior-performance map:} A user reduces a high-dimensional search space to a low-dimensional behavior space by defining dimensions along which behaviors vary. In simulation, the high-dimensional space is then automatically searched to find a high-performing behavior at each point in the low-dimensional behavior space, creating a ``behavior-performance'' map of the performance potential of each location in the low-dimensional space.
In our hexapod robot experiments, the behavior space is six-dimensional: the portion of time that each leg is in contact with the ground. 
The confidence regarding the accuracy of the predicted performance for each behavior in the behavior-performance map is initially low because no tests on the physical robot have been conducted. \textbf{(C \& D) Adaptation Step:} After damage, the robot selects a promising behavior, tests it, updates the predicted performance of that behavior in the behavior-performance map, and sets a high confidence on this performance prediction. The predicted performances of nearby behaviors--and confidence in those predictions--are likely to be similar to the tested behavior and are thus updated accordingly. This select/test/update loop is repeated until a tested behavior on the physical robot performs better than 90\% of the best predicted performance in the behavior-performance map, a value that can decrease with each test (Extended Data Fig.~\ref{fig:scheme_bo}). The algorithm that selects which behavior to test next balances between choosing the behavior with the highest predicted performance and behaviors that are different from those tested so far. Overall, the Intelligent Trial and Error approach presented here rapidly locates which types of behaviors are least affected by the damage to find an effective, compensatory behavior.}
\end{figure*}

The behavior-performance map is created with a novel algorithm and a simulation of the robot, which either can be a standard physics simulator or can be automatically discovered\cite{bongard2006resilient}. The robot's designers only have to describe the dimensions of the space of possible behaviors and a performance measure. For instance, walking gaits could be described by how much each leg is involved in a gait (a behavioral measure) and speed (a performance measure). For grasping, performance could be the amount of surface contact, and it has been demonstrated that 90\% of effective poses for the 21-degree-of-freedom human hand can be captured by a 3-dimensional behavioral space\cite{Santello1998}. To fill in the behavior-performance map, an optimization algorithm simultaneously searches for a high-performing solution at each point in the behavioral space (Fig.~\ref{fig:bo}A,B and Extended Data Fig. \ref{fig:scheme_bo}). This step requires simulating millions of behaviors, but needs to be performed only once per robot design before deployment (Methods).

A low confidence is assigned to the predicted performance of behaviors stored in this behavior-performance map because they have not been tried in reality (Fig.~\ref{fig:bo}B and Extended Data Fig. \ref{fig:scheme_bo}). During the robot's mission, if it senses a performance drop, it selects the most promising behavior from the behavior-performance map, tests it, and measures its performance. The robot subsequently updates its prediction for that behavior and nearby behaviors, assigns high confidence to these predictions (Fig.~\ref{fig:bo}C and Extended Data Fig. \ref{fig:scheme_bo}), and continues the selection/test/update process until it finds a satisfactory compensatory behavior (Fig.~\ref{fig:bo}D and Extended Data Fig. \ref{fig:scheme_bo}).

All of these ideas are technically captured via a Gaussian process model\cite{Rasmussen2006}, which approximates the performance function with already acquired data, and a Bayesian optimization procedure\cite{Mockus2013,Borji2013}, which exploits this model to search for the maximum of the performance function (Methods). The robot selects which behaviors to test by maximizing an information acquisition function that balances exploration (selecting points whose performance is uncertain) and exploitation (selecting points whose performance is expected to be high) (Methods). The selected behavior is tested on the physical robot and the actual performance is recorded. The algorithm updates the expected performance of the tested behavior and lowers the uncertainty about it. These updates are propagated to neighboring solutions in the behavioral space by updating the Gaussian process (Methods). These updated performance and confidence distributions affect which behavior is tested next. This select-test-update loop repeats until the robot finds a behavior whose measured performance is greater than 90\% of the best performance predicted for any behavior in the behavior-performance map (Methods).

We first test our algorithm on a hexapod robot that needs to walk as fast as possible (Fig. \ref{fig:concept}B, D). The robot has 18 motors, an onboard computer, and a depth camera that allows the robot to estimate its walking speed (Supplementary Methods). The gait is parametrized by 36 real-valued parameters that describe the amplitude of oscillation, phase shift, and duty cycle for each joint (Supplementary Methods). The behavior space is 6-dimensional, where each dimension is the proportion of time the \emph{i$^\mathrm{th}$} leg spends in contact with the ground (i.e. the duty factor)\cite{Siciliano2008} (Supplementary Methods).

\begin{figure*}
\floatbox[{\capbeside\thisfloatsetup{capbesideposition={right,top},capbesidewidth=0.25\textwidth}}]{figure}[\FBwidth]
{\caption{\label{fig:results}\textbf{(A) Conditions tested on the physical hexapod robot.} C1: The undamaged robot. C2: One leg is shortened by half. C3: One leg is unpowered. C4: One leg is missing. C5: Two legs are missing. C6: A temporary, makeshift repair to the tip of one leg. \textbf{(B) Performance after adaptation.} Box plots represent Intelligent Trial and Error. The central mark is the median, the edges of the box are the 25\textsuperscript{th} and 75\textsuperscript{th} percentiles, the whiskers extend to the most extreme data points not considered outliers, and outliers are plotted individually. Yellow stars represent the performance of the handmade reference tripod gait (Supplementary Methods). Conditions C1-C6 are tested 5 times each for 8 independently created behavior-performance maps with the ``duty factor'' behavior description (i.e. 40 experiments per damage condition, Supplementary Methods). Damage conditions C1 and C3 are also tested 5 times each for 8 independently created behavior-performance maps with the ``body orientation'' behavior description (Supplementary Methods). \textbf{(C) Time and number of trials required to adapt.} Box plots represent Intelligent Trial and Error.  \textbf{(D) Robotic arm experiment.} The 8-joint, planar robot has to drop a ball into a bin. \textbf{(E) Example conditions tested on the physical robotic arm.} C1: One joint is stuck at 45 degrees. C2: One joint has a permanent 45-degree offset. C3: One broken and one offset joint. A total of 14 conditions were tested (Extended Data Fig. \ref{fig:arm}). \textbf{(F) Time and number of trials required to reach within 5 cm of the bin center.} Each condition is tested with 15 independently created behavior-performance maps.}}
{\includegraphics[width=0.75\textwidth]{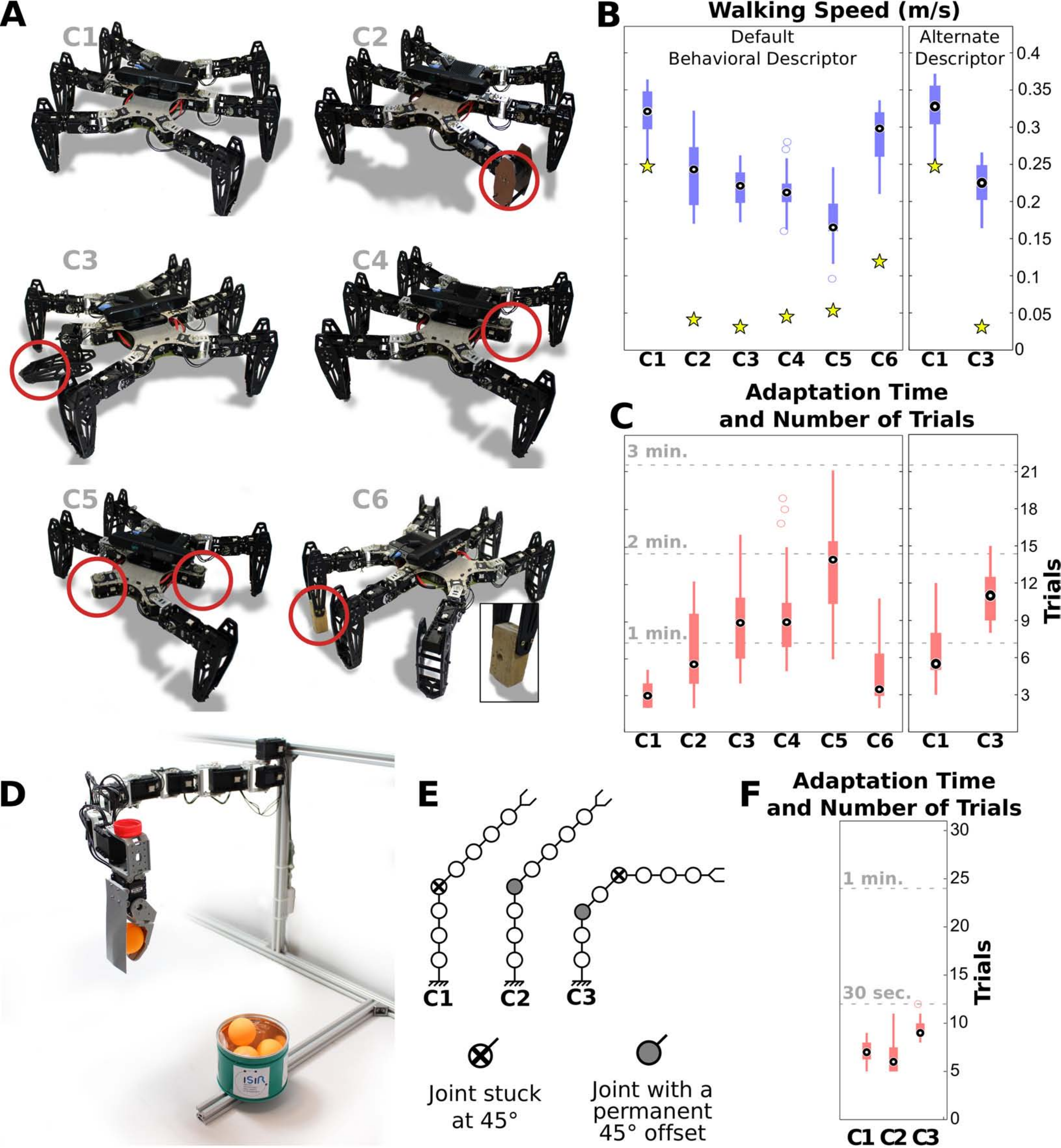}
}
\end{figure*}

The created behavior-performance map contains approximately 13,000 different gaits (Supplementary Video S2 shows examples). We tested our robot in six different conditions: undamaged (Fig.~\ref{fig:results}A:C1), four different structural failures (Fig. \ref{fig:results}A:C2-C5), and a temporary leg repair (Fig. \ref{fig:results}A:C6). We compare the walking speed of resultant gaits with a widely-used, classic, hand-designed tripod gait\cite{Siciliano2008} (Supplementary Methods). For each of the 6 damage conditions, we ran our adaptation step 5 times for each of 8 independently generated behavior-performance maps (with the default ``duty factor'' behavioral description), leading to $6 \times 5 \times 8 = 240$ experiments in total. We also ran our adaptation step 5 times on 8 independently generated behavior-performance maps defined by an alternate behavioral description (``body orientation'', see Supplementary Methods) on two damage conditions~(Fig.~\ref{fig:results}B-C), leading to $2 \times 5 \times 8 = 80$ additional experiments.

When the robot is undamaged (Fig. \ref{fig:results}A:C1), our approach yields dynamic gaits that are $30\%$ faster than the classic reference gait (Fig. \ref{fig:results}B, median 0.32 $m/s$, $5^{th}$ and $95^{th}$ percentiles [0.26; 0.36]  vs. 0.24$m/s$), suggesting that Intelligent Trial and Error is a good search algorithm for automatically producing successful robot behaviors, putting aside damage recovery. In all the damage scenarios, the reference gait is no longer effective (\texttildelow0.04 $m/s$ for the four damage conditions, Fig. \ref{fig:results}B:C2-C5). After Intelligent Trial and Error, the compensatory gaits achieve a reasonably fast speed ($> 0.15 m/s$) and are between $3$ and $7$ times more efficient than the reference gait for that damage condition (in $m/s$, C2: 0.24 [0.18; 0.31] vs. 0.04; C3: 0.22 [0.18; 0.26] vs. 0.03; C4: 0.21 [0.17; 0.26] vs. 0.04; C5: 0.17 [0.12; 0.24] vs. 0.05; C6: 0.3 [0.21; 0.33] vs 0.12). 

These experiments demonstrate that Intelligent Trial and Error allows the robot to both initially learn fast gaits and to reliably recover after physical damage. Additional experiments reveal that these capabilities are substantially faster than state-of-the-art algorithms (Extended Data Fig. \ref{fig:control_exp}), and that Intelligent Trial and Error can help with another major challenge in robotics: adapting to new environments (Extended Data Fig. \ref{fig:exp_slope}). On the undamaged or repaired robot (Fig.~\ref{fig:results}: C6), Intelligent Trial and Error learns a walking gait in less than 30 seconds (Fig. \ref{fig:results}C, undamaged: 24 [16; 41] seconds, 3 [2; 5] physical trials, repaired: 29 [16; 82] seconds, 3.5 [2; 10] trials).
 For the four damage scenarios, the robot adapts in approximately one minute (66 [24; 134] seconds, 8 [3; 16] trials).
 Our results are qualitatively unchanged when using different behavioral characterizations, including randomly choosing 6 descriptors among 63 possibilities (Fig.~\ref{fig:results}B-C and Extended Data Fig. \ref{fig:alt_bd}). Additional experiments show that reducing the high-dimensional parameter space to a low-dimensional behavior space via the behavior-performance map is the key component for intelligent trial and error: standard Bayesian optimization in the original parameter space does not find working controllers (Extended Data Fig. \ref{fig:control_exp}).

\begin{figure*}
\begin{center}
\includegraphics[width=0.8\textwidth]{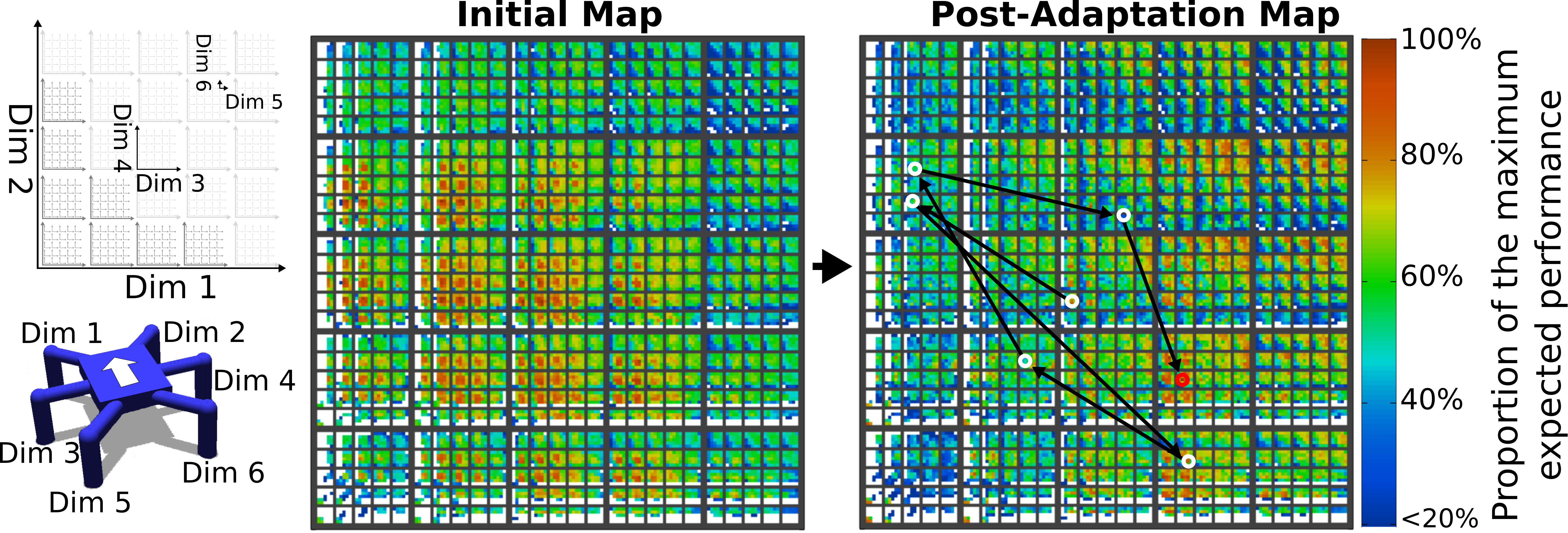}
\end{center}
\caption{\label{fig:matrix}\textbf{An example behavior-performance map.} This map stores high-performing behaviors at each point in a six-dimensional behavior space. Each dimension is the portion of time that each leg is in contact with the ground. The behavioral space is discretized at five values for each dimension (0; 0.25; 0.5; 0.75; 1). Each colored pixel represents the highest-performing behavior discovered during map creation at that point in the behavior space. The matrices visualize the six-dimensional behavioral space in two dimensions according to the legend in the top-left. The behavior-performance map is created with a simulated robot (bottom left) in the Open Dynamics Engine physics simulator (http://www.ode.org). The left matrix is a pre-adaptation map produced by the map creation algorithm. During adaptation, the map is updated as tests are conducted (in this case, in the damage condition where the robot is missing one leg: Fig. \ref{fig:results}A:C4). The right matrix shows the state of the map after a compensatory behavior is discovered. The arrows and white circles represent the order in which behaviors were tested on the physical robot. The red circle is the final, discovered, compensatory behavior. Amongst other areas, high-performing behaviors can be found for the damaged robot in the first two columns of the third dimension. These columns represent behaviors that least use the central-left leg, which is the leg that is missing.}
\end{figure*}

We investigated how the behavior-performance map is updated when the robot loses a leg (Fig. \ref{fig:results}A:C4). Initially the map predicts large areas of high performance. During adaptation, these areas disappear because the behaviors do not work well on the damaged robot. Intelligent Trial and Error quickly identifies one of the few, remaining, high-performance behaviors (Fig.~\ref{fig:matrix} and Extended Data Fig. \ref{fig:archive} and \ref{fig:archive2}).

The same damage recovery approach can be applied to any robot, such as a robotic arm. We tested 14 different damage conditions with a planar, 8-joint robotic arm (Fig. \ref{fig:results}D-F and Extended Data Fig. \ref{fig:arm}). The behavior-performance map's behavioral dimensions are the $x$, $y$ position of the end-effector and the performance measure is minimizing the variance of the 8 specified motor angles (Supplementary Methods). During adaptation, performance is measured as distance to the target. Like with the hexapod robot, our approach discovers a compensatory behavior in less than 2 minutes, usually in less than 30 seconds, and with fewer than 10 trials (Fig. \ref{fig:results}F and Extended Data Fig. \ref{fig:arm}).

While natural animals do not use the specific algorithm we present, there are parallels between Intelligent Trial and Error and animal learning. Like animals, our robot does not have a predefined strategy for how to cope with every possible damage condition: in the face of a new injury, it exploits its intuitions about how its body works to experiment with different behaviors to find what works best. Also like animals\cite{benson2012innovative}, Intelligent Trial and Error allows the quick identification of working behaviors with a few, diverse tests instead of trying behaviors at random or trying small modifications to the best behavior found so far. Additionally, the Bayesian optimization procedure followed by our robot appears similar to the technique employed by humans when they optimize an unknown function\cite{Borji2013}, and there is strong evidence that animal brains learn probability distributions, combine them with prior knowledge, and act as Bayesian optimizers\cite{pouget2013probabilistic, kording2004bayesian}.

An additional parallel is that Intelligent Trial and Error primes the robot for creativity during a motionless period, after which the generated ideas are tested. This process is reminiscent of the finding that some animals start the day with new ideas that they may quickly disregard after experimenting with them\cite{deregnaucourt2005sleep}, and more generally, that sleep improves creativity on cognitive tasks\cite{wagner2004sleep}. 
A final parallel is that the simulator and Gaussian process components of Intelligent Trial and Error are two forms of predictive models, which are known to exist in animals\cite{ito2008control,bongard2006resilient}. All told, we have shown that combining pieces of nature's algorithm, even if differently assembled, moves robots more towards animals by endowing them with the ability to rapidly adapt to unforeseen circumstances. 




\sffamily

\paragraph{Supplementary Information and methods} are appended at the end of this document.

\paragraph{Acknowledgments.} Thanks to Luigi Tedesco, St\'ephane Doncieux, Nicolas Bredeche, Shimon Whiteson, Roberto Calandra, Jacques Droulez, Pierre Bessi\`ere, Florian Lesaint, Charles Thurat, Serena Ivaldi, Jingyu Li, Joost Huizinga, Roby Velez, Henok Mengistu, Tim Clune, and Anh Nguyen for helpful feedback and discussions. 
Thanks to Michael Brashier for the photo of the three-legged dog. 

This work has been funded by the ANR Creadapt project (ANR-12-JS03-0009), the European Research Council (ERC) under the European Union's Horizon 2020 research and innovation programme (grant agreement number 637972), and a Direction G\'ene\'erale de l'Armement (DGA) scholarship to A.C.

\paragraph{Author Contributions} A.C. and J.-B. M designed the study. A.C. and D.T. performed the experiments. A.C., J.-B. M, D.T. and J.C. discussed additional experiments, analyzed the results, and wrote the paper.
\paragraph{Author information} Correspondence and requests for materials should be addressed to J.-B. M.~(email: jean-baptiste.mouret@inria.fr).

\balance
\begin{footnotesize}

\printbibliography[segment=1]
\end{footnotesize}
\end{refsegment}
\newpage
\pagestyle{fancy}

\begin{SI-figure*}
\centering
\includegraphics[width=\linewidth]{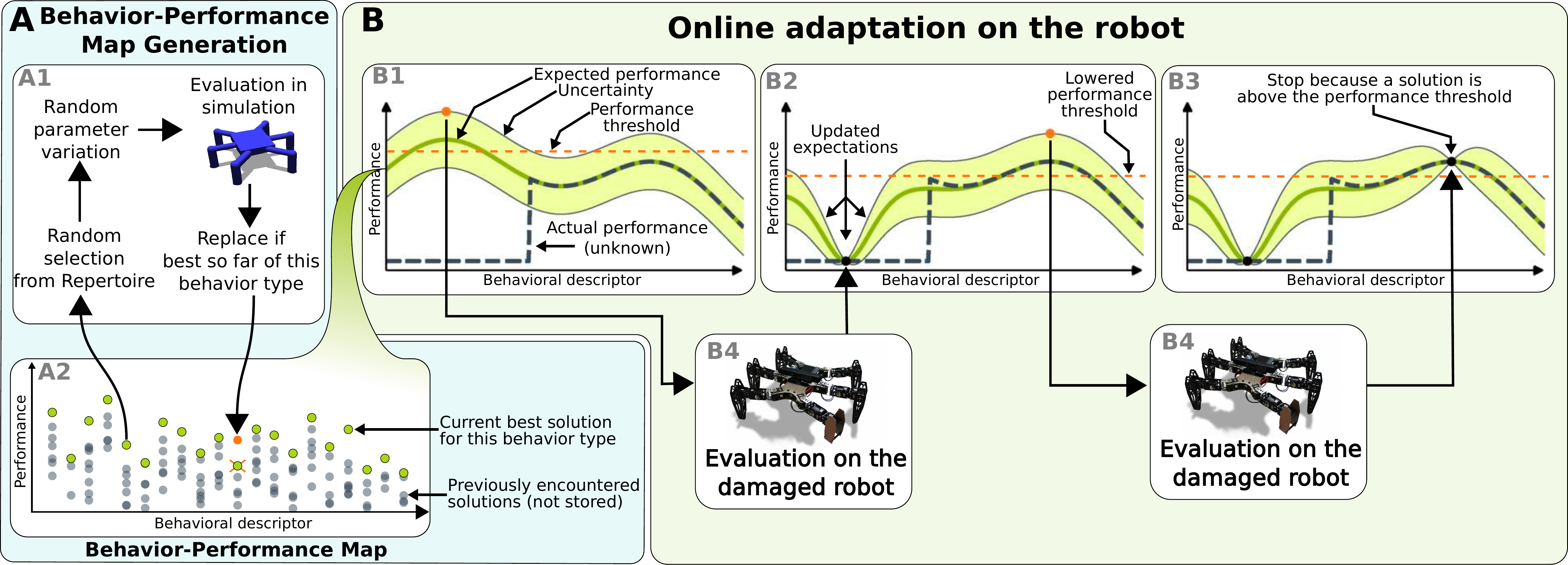}
\caption{\label{fig:scheme_bo} \textbf{An overview of the Intelligent Trial and Error Algorithm.}
  \textbf{(A) Behavior-performance map creation.} After being
  initialized with random controllers, the
  behavioral map (A2), which stores the highest-performing controller found so far of each behavior type, is improved by repeating the process depicted in (A1) until newly generated controllers are rarely good enough to be added to the map (here, after 40 million evaluations). This step, which occurs in simulation, is computationally expensive, but only needs to be performed once per robot (or robot design) prior to deployment. In our experiments, creating one map involved 40 million iterations of (A1), which lasted roughly two weeks on one multi-core computer (Supplementary Methods, section ``Running time''). 
   \textbf{(B) Adaptation.}  (B1) Each behavior from the behavior-performance map has an expected performance based on its performance in simulation (dark green line) and an estimate of uncertainty regarding this predicted performance (light green band). The actual performance on the now-damaged robot (black dashed line) is unknown to the algorithm. A behavior is selected to try on the damaged robot. This selection is made by balancing exploitation---trying behaviors expected to perform well---and exploration---trying behaviors whose performance is uncertain (Methods, section ``acquisition function''). 
   Because all points initially have equal, maximal uncertainty, the first point chosen is that with the highest expected performance. Once this behavior is tested
  on the physical robot (B4), the performance predicted for that behavior is set to its actual performance, the uncertainty regarding that prediction is lowered, and the predictions for, and uncertainties about, nearby controllers are also updated (according to a Gaussian process model, see Methods, section ``kernel function''), the results of which can be seen in (B2). The process is then repeated until performance on the damaged robot is 90\% or greater of the maximum expected performance for any behavior (B3). This performance threshold (orange dashed line) lowers as the maximum expected performance (the highest point on the dark green line) is lowered, which occurs when physical tests on the robot underperform expectations, as occurred in (B2).
  }
\end{SI-figure*}

\begin{SI-figure*}
\centering
\includegraphics[width=0.7\linewidth]{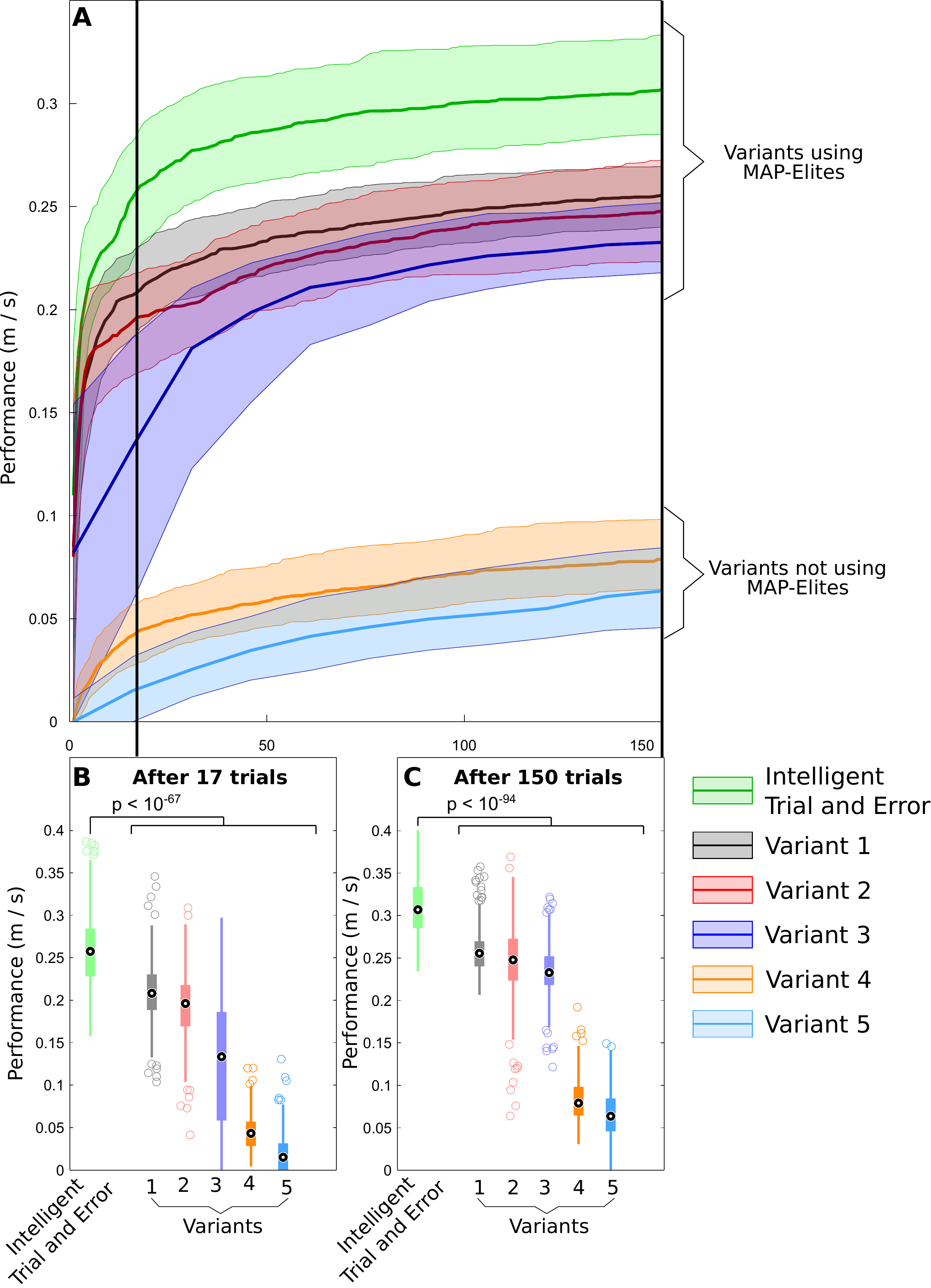}
\bigskip
\begin{small}
\begin{tabular}{lcccc}
\hline
  Variant  & Behavior-performance map & Priors on & Search  & equivalent \\
           & creation            & performance & algorithm & approach\\
  \hline
  Intelligent Trial and Error & MAP-Elites & yes & Bayesian Optimization & -\\
  Variant 1 & MAP-Elites & none & random search & -\\
  Variant 2 & MAP-Elites & none & Bayesian optimization & - \\
  Variant 3 & MAP-Elites & none & policy gradient & - \\
  Variant 4 & none & none & Bayesian optimization & Lizotte et al. (2007)\\
  Variant 5 & none & none & policy gradient & Kohl et al. (2004)\\ 
  \hline
\end{tabular}
\end{small}
\caption{\label{fig:control_exp} \textbf{The contribution of each subcomponent of the Intelligent Trial and Error Algorithm. }
\textbf{(A)} \textbf{Adaptation progress versus the number of robot trials.} The walking speed achieved with Intelligent Trial and Error and several ``knockout'' variants that are missing one of the algorithm's key components. Some variants (4 and 5) correspond to state-of-the-art learning algorithms (policy gradient: Kohl et al. 2004; Bayesian optimization: Lizotte et al. 2007, Tesch et al., 2011, Calandra et al. 2014,). The bold lines represent the medians and the colored areas extend to the 25\textsuperscript{th} and 75\textsuperscript{th} percentiles. \textbf{(B, C)} \textbf{Adaptation performance after 17 and 150 trials.} Shown is the the speed of the compensatory behavior discovered by each algorithm after 17 and 150  evaluations on the robot, respectively. For all panels, data are pooled across six damage conditions (the removal of each of the 6 legs in turn). See Supplementary Experiment S2 for methods and analysis.}
\end{SI-figure*}

\begin{SI-figure*}
\centering
\includegraphics[width=\linewidth]{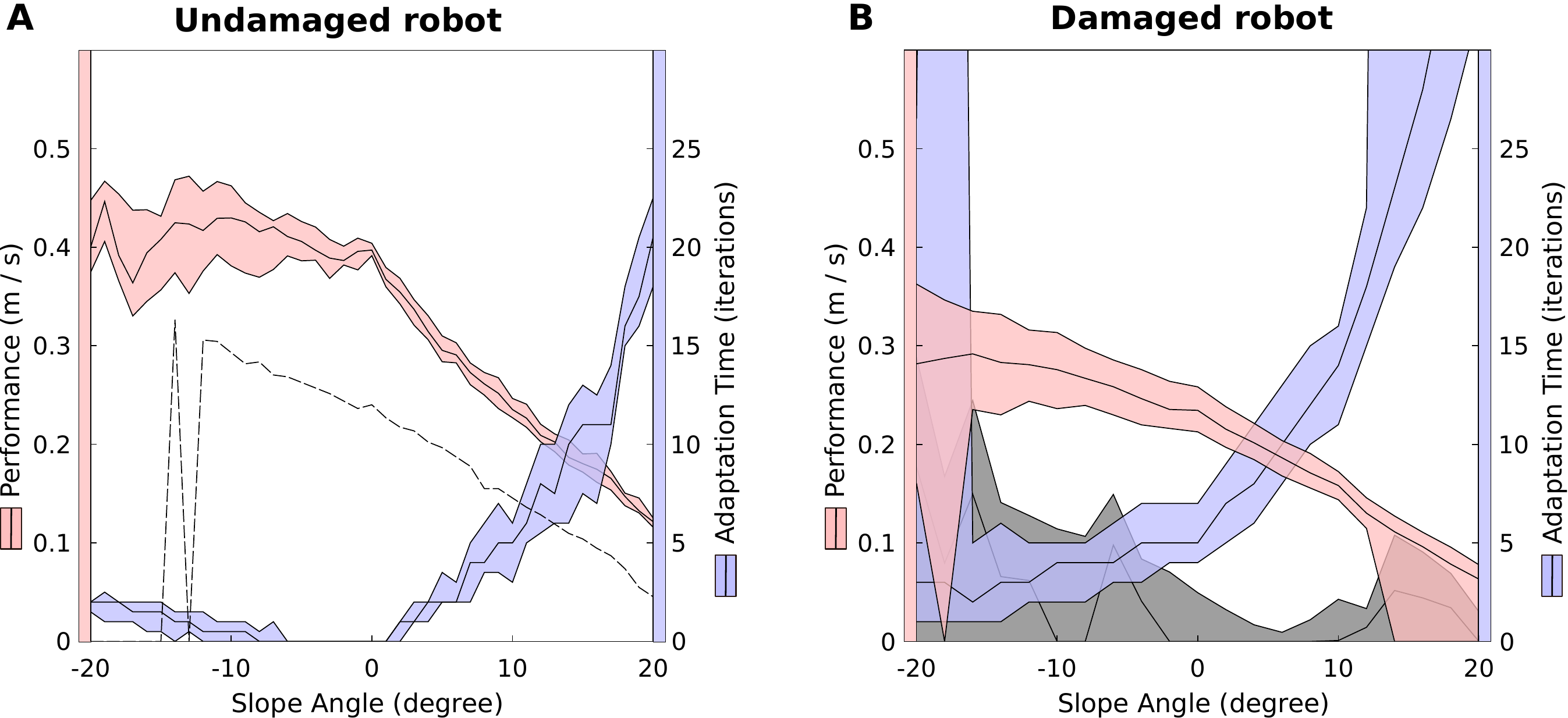}
\caption{\label{fig:exp_slope}\textbf{The Intelligent Trial and Error Algorithm is robust to environmental changes.}
Each plot shows both the performance and required adaptation time for Intelligent Trial and Error when the robot must adapt to walk on terrains of different slopes. \textbf{(A)} \textbf{Adaptation performance on an undamaged robot.} On all slope angles, with very few physical trials, the Intelligent Trial and Error Algorithm (pink shaded region) finds fast gaits that outperform the reference gait (black dotted line). \textbf{(B) Adaptation performance on a damaged robot.} The robot is damaged by having each of the six legs removed in six different damage scenarios. Data are pooled from all six of these damage conditions. The median compensatory behavior found via Intelligent Trial and Error outperforms the median reference controller on all slope angles. The middle, black lines represent medians, while the colored areas extend to the 25\textsuperscript{th} and 75\textsuperscript{th} percentiles. In (A), the black dashed line is the performance of a classic tripod gait for reference. In (B), the reference gait is tried in all six damage conditions and its median (black line) and 25\textsuperscript{th} and 75\textsuperscript{th} percentiles (black colored area) are shown. See Supplementary Experiment S3 for methods and analysis.}
\end{SI-figure*}

\begin{SI-figure*}
\centering
\includegraphics[width=\linewidth]{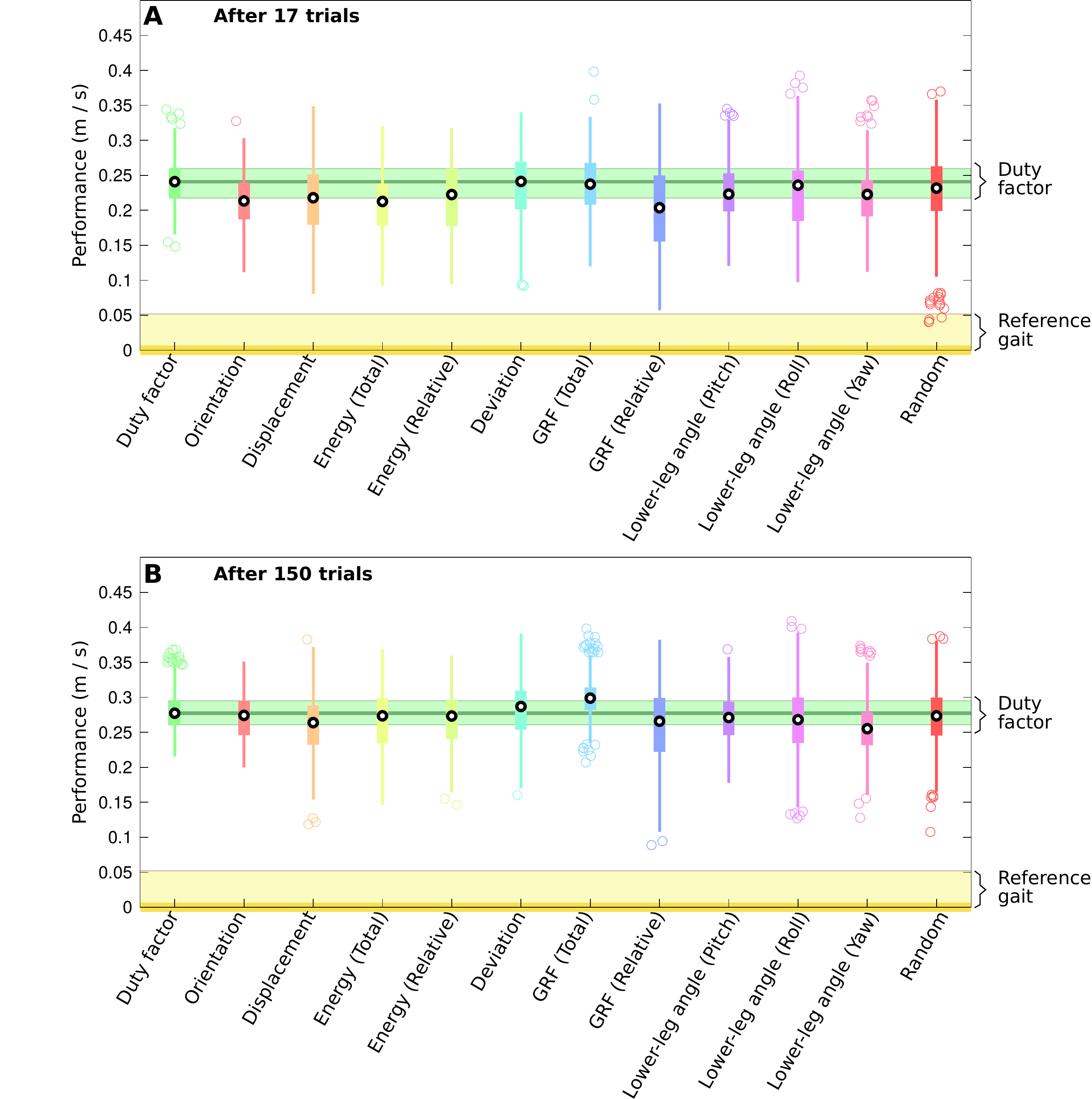}
\caption{\label{fig:alt_bd} \textbf{The Intelligent Trial and Error Algorithm is largely robust to alternate choices of behavior descriptors.} 
\textbf{(A, B)}~The speed of the compensatory behavior discovered by Intelligent Trial and Error for various choices of behavior descriptors. Performance is plotted after 17 and 150 evaluations in panels A and B, respectively. Experiments were performed on a simulated, damaged hexapod. The damaged robot has each of its six legs removed in six different damage scenarios. Data are pooled across all six damage conditions. As described in Supplementary Experiment S5, the evaluated behavior descriptors characterize the following: (i)~Time each leg is in contact with the ground~(\textbf{Duty~factor}); (ii)~Orientation of the robot frame~(\textbf{Orientation}); (iii)~Instantaneous velocity of the robot~(\textbf{Displacement}); (iv)~Energy expended by the robot in walking~(\textbf{Energy~(Total)}, \textbf{Energy~(Relative)}); (v)~Deviation from a straight line~(\textbf{Deviation}); (vi)~Ground reaction force on each leg~(\textbf{GRF (Total)}, \textbf{GRF~(Relative)}); (vii)~The angle of each leg when it touches the ground~(\textbf{Lower-leg angle~(Pitch)}, \textbf{Lower-leg angle~(Roll)}, \textbf{Lower-leg angle~(Yaw)}); and (viii)~A random selection without replacement from subcomponents of all the available behavior descriptors (i-vii)~(\textbf{Random}). For the hand-designed reference gait (yellow) and the compensatory gaits found by the default duty factor behavior descriptor (green), the bold lines represent the medians and the colored areas extend to the 25\textsuperscript{th} and 75\textsuperscript{th} percentiles of the data. For the other treatments, including the duty factor treatment, black circles represent the median, the colored area extends to the 25\textsuperscript{th} and 75\textsuperscript{th} percentiles of the data, and the colored circles are outliers. See Supplementary Experiment S5 for methods and analysis.
}
\end{SI-figure*}

\begin{SI-figure*}
\centering
\includegraphics[width=\linewidth]{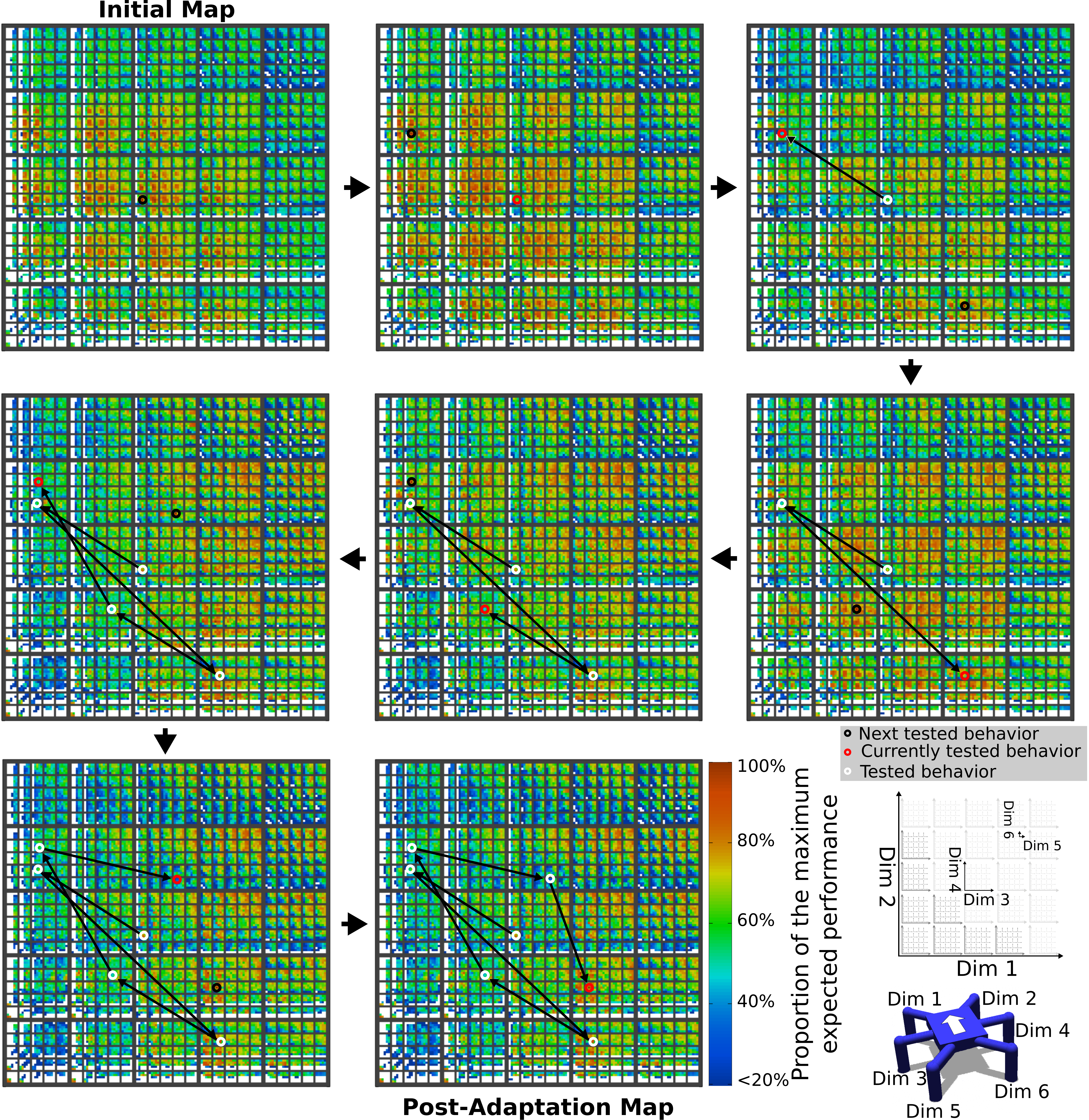}
\caption{\label{fig:archive} \textbf{How the behavior performance map is explored to discover a compensatory behavior (normalized each iteration to highlight the \emph{range} of remaining performance predictions).} 
Colors represent the performance prediction for each point in the map relative to the highest performing prediction in the map at that step of the process. 
A black circle indicates the next behavior to be tested on the physical robot. A red circle indicates the behavior that was just tested (note that the performance predictions surrounding it have changed versus the previous panel). Arrows reveal the order that points have been explored. The red circle in the last map is the final, selected, compensatory behavior. In this scenario, the robot loses leg number 3. The six dimensional space is visualized according to the inset legend.}
\end{SI-figure*}
\begin{SI-figure*}
\centering
\includegraphics[width=\linewidth]{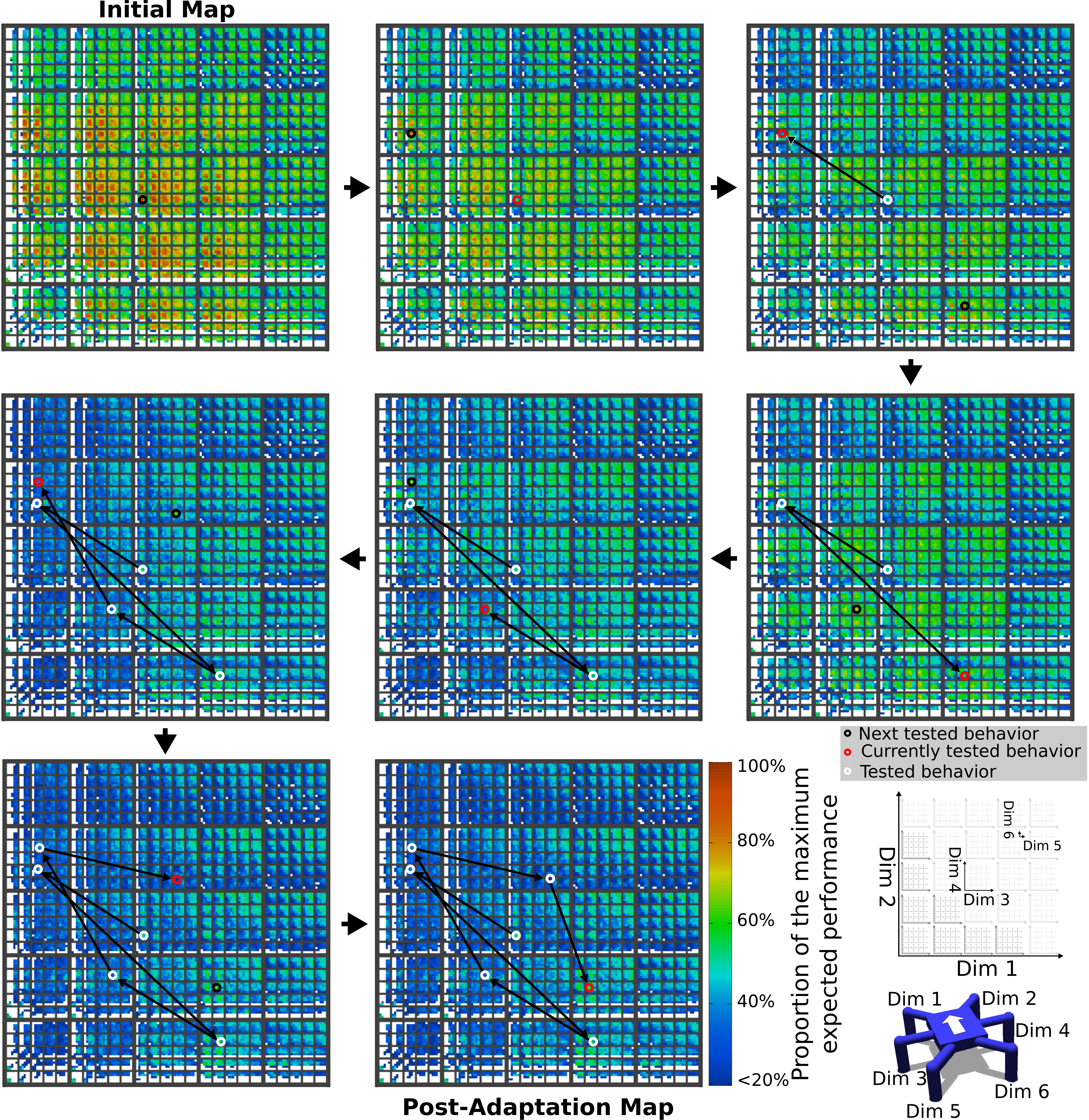}
\caption{\label{fig:archive2} \textbf{How the behavior performance map is explored to discover a compensatory behavior (non-normalized to highlight that performance predictions decrease as it is discovered that predictions from the simulated, undamaged robot do not work well on the damaged robot).} 
Colors represent the performance prediction for each point in the map relative to the highest performing prediction in the \emph{first} map. 
A black circle indicates the next behavior to be tested on the physical robot. A red circle indicates the behavior that was just tested (note that the performance predictions surrounding it have changed versus the previous panel). Arrows reveal the order that points have been explored. The red circle in the last map in the sequence is the final, selected, compensatory behavior. In this scenario, the robot loses leg number 3. The six dimensional space is visualized according to the inset legend. The data visualized in this figure are identical to those in the previous figure: the difference is simply whether the data are renormalized for each new map in the sequence.}
\end{SI-figure*}

\begin{SI-figure*}
\centering
\includegraphics[width=\linewidth]{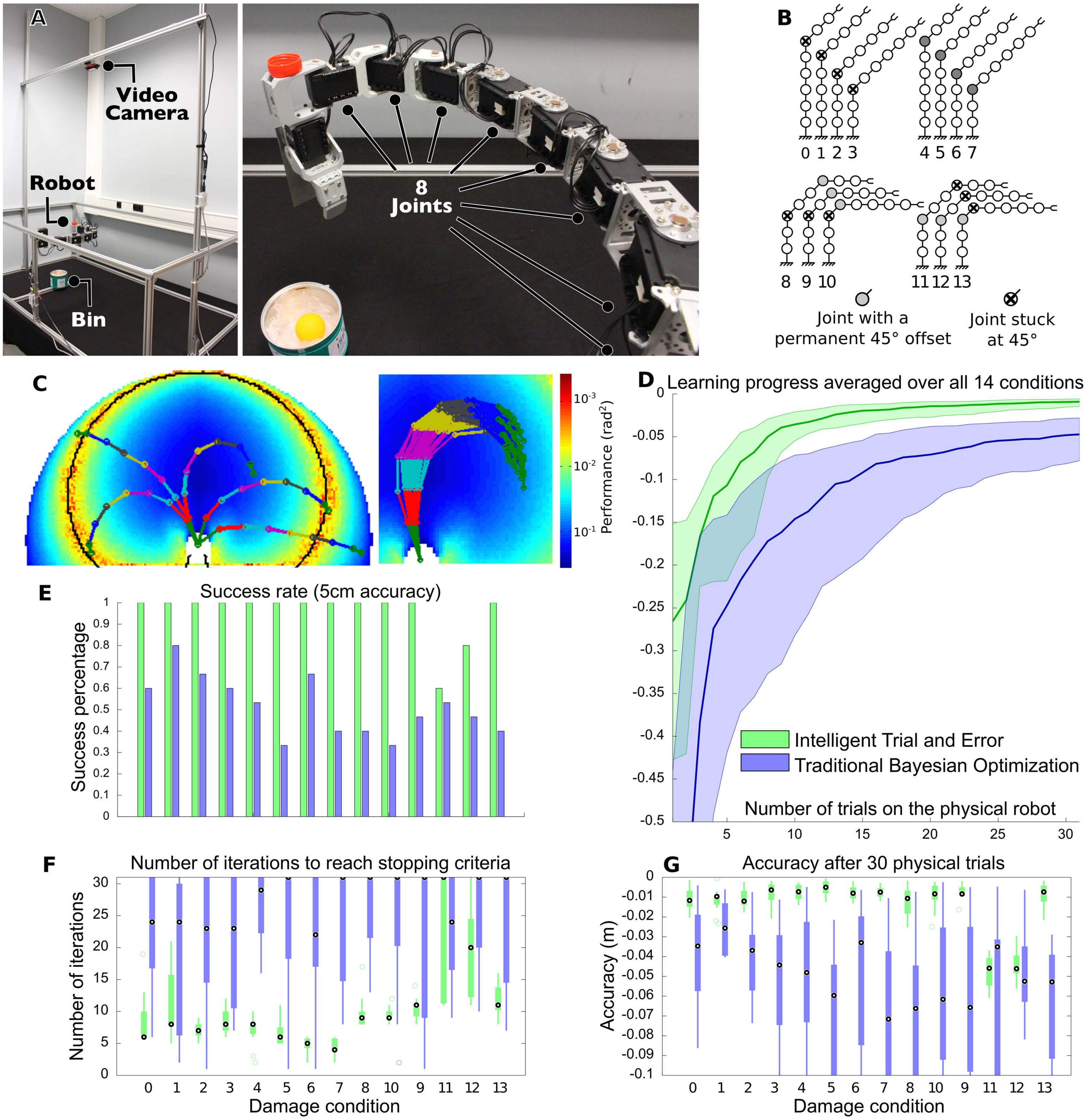}
\caption{\label{fig:arm} \textbf{Intelligent Trial and Error works on a completely different type of robot: supplementary data from the robotic arm experiment.} \textbf{(A) The robotic arm experimental setup.} \textbf{(B) Tested damage conditions}. \textbf{(C) Example of behavior performance maps (colormaps) and behaviors (overlaid arm configurations) obtained with MAP-Elites.} Left: A typical behavior-performance map produced by MAP-Elites with 5 example behaviors, where a behavior is described by the angle of each of the 8 joints. The color of each point is a function of its performance, which is defined as having low variance in the joint angles (i.e. a zigzag arm is lower performing than a straighter arm that reaches the same point). Right: Neighboring points in the map tend to have similar behaviors, thanks to the performance function, which would penalize more jagged ways of reaching those points. That neighbors have similar behaviors justifies updating predictions about the performance of nearby behaviors after a testing a single behavior on the real (damaged) robot. \textbf{(D) Performance vs. trial number for Intelligent Trial and Error and traditional Bayesian optimization.} The experiment was conducted on the physical robot, with 15 independent replications for each of the 14 damage conditions. Performance is pooled from all of these $14\times15=210$ experiments. \textbf{(E) Success for each damage condition.} Shown is the success rate for the 15 replications for each damage condition, defined as the percentage of replicates in which the robot reaches within 5 cm of the bin center. \textbf{(F) Trials required to adapt.} Shown is the number of iterations required to reach within 5 cm of the basket center. \textbf{(G) Accuracy after 30 physical trials.} Performance after 30 physical trials for each damage condition (with the stopping criterion disabled). See Supplementary Experiment S1 for methods and analysis. }
\end{SI-figure*}

\section{Methods}
\begin{refsegment}
\subsection*{Notations}
\label{alg:global}
\begin{itemize}
  \item $\mathbf{c}$: Parameters of a controller (vector)
  \item $\mathbf{x}$: A location in a discrete behavioral space (i.e. a type of behavior) (vector)
  \item $\mathbf{\chi}$: A location in a discrete behavioral space that has been tested on the physical robot (vector)
  \item $\mathcal{P}$: Behavior-performance map (stores performance) (associative table)
  \item $\mathcal{C}$: Behavior-performance map (stores controllers) (associative table)
  \item $\mathcal{P}(\mathbf{x})$: Max performance yet encountered at $\mathbf{x}$ (scalar)
  \item $\mathcal{C}(\mathbf{x})$: Controller currently stored in $\mathbf{x}$ (vector)
  \item $\mathbf{\chi_{1:t}}$: All previously tested behavioral descriptors at time $t$ (vector of vectors)
  \item $\mathbf{P_{1:t}}$: Performance in reality of all the candidate solutions tested on the robot up to time $t$ (vector)
  \item $\mathcal{P}(\mathbf{\chi_{1:t}})$: Performance in the behavior-performance map for all the candidate solutions tested on the robot up to time $t$ (vector)
  \item $f()$: Performance function (unknown by the algorithm) (function)
  \item $\sigma_{noise}^2$: Observation noise (a user-specified parameter) (scalar)
  \item $k(\mathbf{x},\mathbf{x})$: Kernel function (see section ``kernel function'') (function)
  \item $\mathbf{K}$: Kernel matrix (matrix)
  \item $\mathbf{k}$: Kernel vector $[k(\mathbf{x}, \chi_1), k(\mathbf{x}, \chi_2),...,k(\mathbf{x}, \chi_t)]$ (vector)
  \item $\mu_{t}(\mathbf{x})$: Predicted performance for $\mathbf{x}$ (i.e. the mean of the Gaussian process) (function)
  \item $\sigma_{t}^2(\mathbf{x})$: Standard deviation for $\mathbf{x}$ in the Gaussian process (function)
\end{itemize}

\subsection{Intelligent Trial and Error algorithm (IT\&E)}
The Intelligent Trial and Error Algorithm consists of two major steps (Extended Data Fig. \ref{fig:scheme_bo}): the behavior-performance map creation step and the adaptation step (while here we focus on damage recovery, Intelligent Trial and Error can search for any type of required adaptation, such as learning an initial gait for an undamaged robot, adapting to new environments, etc.). The behavior-performance map creation step is accomplished via a new algorithm introduced in this paper called multi-dimensional archive of phenotypic elites (MAP-Elites), which is explained in the next section. The adaptation step is accomplished via a second new algorithm introduced in this paper called the map-based Bayesian optimization algorithm (M-BOA), which is explained in the ``Adaptation Step'' section below.




\subsection{Behavior-performance map creation (via the MAP-Elites algorithm)}
\label{alg:archive}

The behavior-performance map is created by a new algorithm we introduce in this paper called the multi-dimensional archive of phenotypic elites (MAP-Elites) algorithm. MAP-Elites searches for the highest-performing solution for each point in a user-defined space: the user chooses the dimensions of the space that they are interested in seeing variation in. For example, when designing robots, the user may be interested in seeing the highest-performing solution at each point in a two-dimensional space where one axis is the weight of the robot and the other axis is the height of the robot. Alternatively, a user may wish to see weight vs. cost, or see solutions throughout a 3D space of weight vs. cost vs. height. Any dimension that can vary could be chosen by the user. There is no limit on the number of dimensions that can be chosen, although it becomes computationally more expensive to fill the behavior-performance map and store it as the number of dimensions increases. It also becomes more difficult to visualize the results. We refer to this user-defined space as the ``behavior space'', because usually the dimensions of variation measure behavioral characteristics. Note that the behavioral space can refer to other aspects of the solution (as in this example, where the dimensions of variation are physical properties of a robot such as its height and weight). 

If the behavior descriptors and the parameters of the controller are the same (i.e. if there is only one possible solution/genome/parameter set/policy/description for each location in the behavioral space), then creating the behavior-performance map is straightforward: one simply needs to simulate the solution at each location in the behavior space and record the performance. However, if it is not known a priori how to produce a controller/parameter set/description that will end up in a specific location in the behavior space (i.e. if the parameter space is of higher dimension than the behavioral space: e.g., in our example, if there are many different robot designs of a specific weight, height, and cost, or if it is unknown how to make a description that will produce a robot with a specific weight, height, and cost), then MAP-Elites is beneficial. It will efficiently search for the highest-performing solution at each point of the low-dimensional behavioral space. It is more efficient than a random sampling of the search space because high-performing solutions are often similar in many ways, such that randomly altering a high-performing solution of one type can produce a high-performing solution of a different type (see Extended Data Fig. \ref{fig:map-elites} and Supplementary Experiment S4). For this reason, searching for high-performing solutions of all types simultaneously is much quicker than separately searching for each type. For example,  to generate a lightweight, high-performing robot design, it tends to be more effective and efficient to modify an existing design of a light robot rather than randomly generate new designs from scratch or launch a separate search process for each new type of design.

MAP-Elites begins by generating a set of random candidate solutions. It then evaluates the performance of each solution and records where that solution is located in the behavior space (e.g. if the dimensions of the behavior space are the height and weight, it records the height and weight of each robot in addition to its performance). For each solution, if its performance is better than the current solution at that location in the behavior-performance map, then it is added to the behavior-performance map, replacing the solution in that location. In other words, it is only kept if it is the best of that type of solution, where ``type'' is defined as a location in the behavior space. There is thus only one solution kept at each location in the behavior space (keeping more could be beneficial, but for computational reasons we only keep one). If no solution is present in the behavior-performance map at that location, then the newly generated candidate solution is added at that location. 

Once this initialization step is finished, Map-Elites enters a loop that is similar to stochastic, population-based, optimization algorithms, such as evolutionary algorithms\cite{Eiben2003}: the solutions that are in the behavior-performance map form a population that is improved by random variation and selection. In each generation, the algorithm picks a solution at random via a uniform distribution, meaning that each solution has an equal chance of being chosen. A copy of the selected solution is then randomly mutated to change it in some way, its performance is evaluated, its location in the behavioral space is determined, and it is kept if it outperforms the current occupant at that point in the behavior space (note that mutated solutions may end up in different behavior space locations than their ``parents''). This process is repeated until a stopping criterion is met (e.g. after a fixed amount of time has expired). In our experiments, we stopped each MAP-Elites run after 40 million iterations. Because MAP-Elites is a stochastic search process, each resultant behavior-performance map can be different, both in terms of the number of locations in the behavioral space for which a candidate is found, and in terms of the performance of the candidate in each location.


The pseudo-code of the algorithm is available in Supplementary Figure \ref{fig:pseudo-code}. More details and experiments about MAP-Elites are available in (Mouret and Clune, 2015)\cite{mouret2015illuminating}. 

\subsection{Adaptation step (via M-BOA: the map-based Bayesian optimization algorithm)}
\label{alg:rboa}

The adaptation step is accomplished via a Bayesian optimization algorithm seeded with a behavior-performance map. We call this approach a map-based Bayesian optimization algorithm, or M-BOA.

Bayesian optimization is a model-based, black-box optimization algorithm that is tailored for very expensive objective functions (a.k.a. cost functions)\cite{lizotte2007automatic,brochu2010tutorial,Mockus2013,Snoek2012,Griffith2009,Borji2013}. As a black-box optimization algorithm, Bayesian optimization searches for the maximum of an unknown objective function from which samples can be obtained (e.g., by measuring the performance of a robot). Like all model-based optimization algorithms (e.g. surrogate-based algorithms\cite{booker1999rigorous,forrester2009recent,jin2011surrogate}, kriging\cite{simpson1998comparison}, or DACE\cite{jones1998efficient,sacks1989design}), Bayesian optimization creates a model of the objective function with a regression method, uses this model to select the next point to acquire, then updates the model, etc. It is called \emph{Bayesian} because, in its general formulation\cite{Mockus2013}, this algorithm chooses the next point by computing a posterior distribution of the objective function using the likelihood of the data already acquired and a prior on the type of function.

Here we use Gaussian process regression to find a model\cite{Rasmussen2006}, which is a common choice for Bayesian optimization\cite{calandra2014experimental, Griffith2009, brochu2010tutorial, lizotte2007automatic}. Gaussian processes are particularly interesting for regression because they not only model the cost function, but also the uncertainty associated with each prediction. For a cost function $f$, usually unknown, a Gaussian process defines the probability distribution of the possible values $f(\mathbf{x})$ for each point $\mathbf{x}$. These probability distributions are Gaussian, and are therefore defined by a mean ($\mu$) and a standard deviation ($\sigma$). However, $\mu$ and $\sigma$ can be different for each $\mathbf{x}$; we therefore define a probability distribution \emph{over functions}:
 \begin{equation}
P(f(\mathbf{x})|\mathbf{x}) = \mathcal{N}(\mu(\mathbf{x}), \sigma^2(\mathbf{x}))
\end{equation}
where $\mathcal{N}$ denotes the standard normal distribution.

To estimate $\mu(\mathbf{x})$ and $\sigma(\mathbf{x})$, we need to fit the Gaussian process to the data. To do so, we assume that each observation $f(\mathbf{\chi})$ is a sample from a normal distribution. If we have a data set made of several observations, that is, $f(\mathbf{\chi}_1), f(\mathbf{\chi}_2), ..., f(\mathbf{\chi}_t)$, then the vector $\left[f(\mathbf{\chi}_1), f(\mathbf{\chi}_2), ..., f(\mathbf{\chi}_t)\right]$ is a sample from a \emph{multivariate} normal distribution, which is defined by a mean vector and a covariance matrix. A Gaussian process is therefore a generalization of a $n$-variate normal distribution, where $n$ is the number of observations. The covariance matrix is what relates one observation to another: two observations that correspond to nearby values of $\chi_1$ and $\chi_2$ are likely to be correlated (this is a prior assumption based on the fact that functions tend to be smooth, and is injected into the algorithm via a prior on the likelihood of functions), two observations that correspond to distant values of $\chi_1$ and $\chi_2$ should not influence each other (i.e. their distributions are not correlated). Put differently, the covariance matrix represents that distant samples are almost uncorrelated and nearby samples are strongly correlated. This covariance matrix is defined via a \emph{kernel function}, called $k(\chi_1, \chi_2)$, which is usually based on the Euclidean distance between $\chi_1$ and $\chi_2$ (see the ``kernel function'' sub-section below).

Given a set of observations $\mathbf{P}_{1:t}=f(\mathbf{\chi}_{1:t})$ and a sampling noise $\sigma^2_{noise}$(which is a user-specified parameter), the Gaussian process is computed as follows\cite{brochu2010tutorial,Rasmussen2006}:

\begin{equation}\label{eq:GP}
\begin{gathered}
 P(f(\mathbf{x})|\mathbf{P}_{1:t},\mathbf{x}) = \mathcal{N}(\mu_{t}(\mathbf{x}), \sigma_{t}^2(\mathbf{x}))\\
\begin{array}{l}
 \mathrm{where:}\\
 \mu_{t}(\mathbf{x})= \mathbf{k}^\intercal\mathbf{K}^{-1}\mathbf{P}_{1:t}\\
 \sigma_{t}^2(\mathbf{x})=k(\mathbf{x},\mathbf{x}) - \mathbf{k}^\intercal\mathbf{K}^{-1}\mathbf{k}\\
 \mathbf{K}=\left[ \begin{array}{ c c c}
    k(\mathbf{\chi}_1,\mathbf{\chi}_1) &\cdots & k(\mathbf{\chi}_1,\mathbf{\chi}_{t}) \\
    \vdots   &  \ddots &  \vdots  \\
    k(\mathbf{\chi}_{t},\mathbf{\chi}_1) &  \cdots &  k(\mathbf{\chi}_{t},\mathbf{\chi}_{t})\end{array} \right]
+ \sigma_{noise}^2I\\
 \mathbf{k}=\left[ \begin{array}{ c c c c }k(\mathbf{x},\mathbf{\chi}_1) & k(\mathbf{x},\mathbf{\chi}_2) & \cdots & k(\mathbf{x},\mathbf{\chi}_{t}) \end{array} \right]
 \end{array}
\end{gathered}
\end{equation}
 
Our implementation of Bayesian optimization uses this Gaussian process model to search for the maximum of the objective function $f(\mathbf{x})$, $f(\mathbf{x})$ being unknown. It selects the next $\chi$ to test by selecting the maximum of the \emph{acquisition function}, which balances exploration -- improving the model in the less explored parts of the search space -- and exploitation -- favoring parts that the models predicts as promising. Here, we use the ``Upper Confidence Bound'' acquisition function (see the ``information acquisition function'' section below). Once the observation is made, the algorithm updates the Gaussian process to take the new data into account. In classic Bayesian optimization, the Gaussian process is initialized with a constant mean because it is assumed that all the points of the search space are equally likely to be good. The model is then progressively refined after each observation.

The key concept of the map-based Bayesian optimization algorithm (M-BOA) is to use the output of MAP-Elites as a prior for the Bayesian optimization algorithm: thanks to the simulations, we expect some behaviors to perform better than others on the robot. To incorporate this idea into the Bayesian optimization, M-BOA models the \emph{difference} between the prediction of the behavior-performance map and the actual performance on the real robot, instead of directly modeling the objective function. This idea is incorporated into the Gaussian process by modifying the update equation for the mean function ($\mu_t(\mathbf{x})$, equation~\ref{eq:GP}): 

\begin{equation}
\mu_{t}(\mathbf{x})= \mathcal{P}(\mathbf{x}) + \mathbf{k}^\intercal\mathbf{K}^{-1}(\mathbf{P}_{1:t}-\mathcal{P}(\mathbf{\chi}_{1:t}))
\label{eq:rboa}
\end{equation}
where $\mathcal{P}(\mathbf{x})$ is the performance of $\mathbf{x}$ according to the simulation and $\mathcal{P}(\mathbf{\chi}_{1:t})$ is the performance of all the previous observations, also according to the simulation. Replacing $\mathbf{P}_{1:t}$ (eq. \ref{eq:GP}) by $\mathbf{P}_{1:t}-\mathcal{P}(\mathbf{\chi}_{1:t})$ (eq. \ref{eq:rboa}) means that the Gaussian process models the difference between the actual performance $\mathbf{P}_{1:t}$ and the performance from the behavior-performance map $\mathcal{P}(\mathbf{\chi}_{1:t})$. The term $\mathcal{P}(\mathbf{x})$ is the prediction of the behavior-performance map. M-BOA therefore starts with the prediction from the behavior-performance map and corrects it with the Gaussian process.

The pseudo-code of the algorithm is available in Supplementary Figure \ref{fig:pseudo-code}.
\begin{SI-figure*}
\begin{algorithmic}
\label{alg:itel}
\Procedure{Intelligent Trial and Error Algorithm (IT\&E)}{}
\State Before the mission:
\State \Call{\indent Create behavior-performance map (via the MAP-Elites algorithm in simulation)}{}

\While{In mission}
\If{Significant performance fall}
\State \Call{Adaptation Step (via M-BOA algorithm)}{}
\EndIf
\EndWhile
\EndProcedure
\end{algorithmic}

\begin{algorithmic}
\Procedure{MAP-Elites Algorithm}{}
\State $(\mathcal{P} \leftarrow \emptyset, \mathcal{C} \leftarrow \emptyset)$\Comment{\emph{Creation of an empty behavior-performance map (empty N-dimensional grid).}}
\For{iter $  = 1\to I$} \Comment{\emph{Repeat during $I$ iterations (here we choose I = 40 million iterations).}}
\If{iter $< 400$} 
  \State $\mathbf{c'}\leftarrow $ random\_controller()   \Comment{\emph{The first 400 controllers are generated randomly.}}
\Else \Comment{\emph{The next controllers are generated using the map.}}
  \State $\mathbf{c}\leftarrow $ random\_selection($\mathcal{C}$) \Comment{\emph{Randomly select a controller $c$ in the map.}}
  \State $\mathbf{c'}\leftarrow $ random\_variation($\mathbf{c}$) \Comment{\emph{Create a randomly modified copy of $c$.}}
\EndIf
\State $\mathbf{x'}\leftarrow $behavioral\_descriptor(simu($\mathbf{c'}$)) \Comment{\emph{Simulate the controller and record its behavioral descriptor.}}
\State $p'\leftarrow $performance(simu($\mathbf{c'}$)) \Comment{\emph{Record its performance.}}
\If{$\mathcal{P}(\mathbf{x'})= \emptyset$ or $\mathcal{P}(\mathbf{x'})<p'$}\Comment{\emph{If the cell is empty or if $p'$ is better than the current stored performance.}}
\State $\mathcal{P}(\mathbf{x'})\leftarrow p'$ \Comment{\emph{Store the performance of $\mathbf{c'}$ in the behavior-performance map according}}
\State \Comment{\emph{to its behavioral descriptor $\mathbf{x'}$.}}
\State $\mathcal{C}(\mathbf{x'})\leftarrow \mathbf{c'}$ \Comment{\emph{Associate the controller with its behavioral descriptor.}}
\EndIf
\EndFor
\State \Return behavior-performance map ($\mathcal{P}$ and $\mathcal{C}$)
\EndProcedure
\end{algorithmic}

\begin{algorithmic}
\label{alg:bo}
\Procedure{M-BOA (Map-based Bayesian Optimization Algorithm)}{}
\State $\forall \mathbf{x} \in \textrm{map}$:\Comment{\emph{Initialisation.}}
\State \hspace{\algorithmicindent}$P(f(\mathbf{x})|\mathbf{x}) = \mathcal{N}(\mu_0(\mathbf{x}), \sigma_0^2(\mathbf{x}))$\Comment{\emph{Definition of the Gaussian Process.}}
\State \hspace{\algorithmicindent} $\textrm{where}$
\State \hspace{\algorithmicindent} $\mu_0(\mathbf{x})= \mathcal{P}(\mathbf{x}) $\Comment{\emph{Initialize the mean prior from the map.}}
\State \hspace{\algorithmicindent} $\sigma_0^2(\mathbf{x})=k(\mathbf{x},\mathbf{x})$\Comment{\emph{Initialize the variance prior (in the common case, $k(\mathbf{x},\mathbf{x})=1$).}}
\While{$\max(\mathbf{P_{1:t}} )<\alpha \max(\mu_t\mathbf{(x)})$}\Comment{\emph{Iteration loop.}}
\State$\mathbf{\chi}_{t+1}\leftarrow  \operatorname*{arg\,max}_\mathbf{x} (\mu_{t}(\mathbf{x})+ \kappa\sigma_t(\mathbf{x}))$ \Comment{\emph{Select next test (argmax of acquisition function).}}
\State $P_{t+1}\leftarrow $ performance(physical\_robot($\mathcal{C}(\mathbf\chi_{t+1})$)).\Comment{\emph{Evaluation of ~$\mathbf{{x}_{t+1}}$ on the physical robot.}}
\State $P(f(\mathbf{x})|\mathbf{P}_{1:t+1},\mathbf{x}) = \mathcal{N}(\mu_{t+1}(\mathbf{x}), \sigma_{t+1}^2(\mathbf{x}))$ \Comment{\emph{Update the Gaussian Process.}}
\State $\textrm{where}$
\State $\mu_{t+1}(\mathbf{x})= \mathcal{P}(\mathbf{x}) + \mathbf{k}^\intercal\mathbf{K}^{-1}(\mathbf{P}_{1:t+1}-\mathcal{P}(\mathbf{\chi}_{1:t+1}))$\Comment{\emph{Update the mean.}}
\State $\sigma_{t+1}^2(\mathbf{x})=k(\mathbf{x},\mathbf{x}) - \mathbf{k}^\intercal\mathbf{K}^{-1}\mathbf{k}$\Comment{\emph{Update the variance.}}

\State $\mathbf{K}=\left[ \begin{array}{ c c c}
    k(\mathbf{\chi}_1,\mathbf{\chi}_1) &\cdots & k(\mathbf{\chi}_1,\mathbf{\chi}_{t+1}) \\
    \vdots   &  \ddots &  \vdots  \\
    k(\mathbf{\chi}_{t+1},\mathbf{\chi}_1) &  \cdots &  k(\mathbf{\chi}_{t+1},\mathbf{\chi}_{t+1})\end{array} \right]
+ \sigma_{noise}^2I$\Comment{\emph{Compute the observations' correlation matrix.}}
\State $\mathbf{k}=\left[ \begin{array}{ c c c c }k(\mathbf{x},\mathbf{\chi}_1) & k(\mathbf{x},\mathbf{\chi}_2) & \cdots & k(\mathbf{x},\mathbf{\chi}_{t+1}) \end{array} \right]$\Comment{\emph{Compute the $\mathbf{x}$ vs. observation correlation vector.}}

\EndWhile
\EndProcedure
\end{algorithmic}

\caption{\label{fig:pseudo-code} \textbf{Pseudo-code for the Intelligent Trial and Error Algorithm, the MAP-Elites algorithm, and the Map-based Bayesian Optimization Algorithm (M-BOA)}. Notations are described at the beginning of the methods section.}
\end{SI-figure*}

\paragraph{Kernel function}\label{sec:kernel}
The kernel function is the covariance function of the Gaussian
process. It defines the influence of a controller's performance (on the
physical robot)	on the performance and confidence estimations of
not-yet-tested controllers in the behavior-performance map that are nearby in
behavior space to the tested controller (Extended Data Fig. \ref{fig:param_rho}a).

The Squared Exponential covariance function and the Mat\'{e}rn kernel are the most common kernels for Gaussian processes\cite{brochu2010tutorial,Snoek2012,Rasmussen2006}. Both kernels are variants of the ``bell curve''. Here we chose the Mat\'{e}rn kernel because it is more general (it includes the Squared Exponential function as a special case) and because it allows us to control not only the distance at which effects become nearly zero (as a function of parameter $\rho$, Extended Data Fig. \ref{fig:param_rho}a), but also the rate at which distance effects decrease (as a function of parameter $\nu$). 

The Mat\'{e}rn kernel function is computed as follows\cite{matern1960spatial, stein1999interpolation} (with $\nu=5/2$):
\begin{equation}
\begin{array}{l}
k(\mathbf{x}_1,\mathbf{x}_2)=\left(1+ \frac{\sqrt{5}d(\mathbf{x}_1,\mathbf{x}_2)}{\rho}+\frac{5d(\mathbf{x}_1,\mathbf{x}_2)^2}{3\rho^2}\right)\exp\left(-\frac{\sqrt{5}d(\mathbf{x}_1,\mathbf{x}_2)}{\rho}\right)\\
\textrm{where }d(\mathbf{x}_1,\mathbf{x}_2) \textrm{ is the Euclidean distance in behavior space.}
\end{array}
\end{equation}

Because the model update step directly depends on $\rho$, it is one of
the most critical parameters of the Intelligent Trial and Error
Algorithm. We selected its value after extensive experiments in simulation (Extended Data Fig. \ref{fig:param_rho} and section \ref{sec:rho}). 

\paragraph{Information acquisition function}\label{sec:acquisition}
The information acquisition function selects the next solution that
will be evaluated on the physical robot. The selection is made by
finding the solution that maximizes the acquisition function. This
step is another optimization problem, but
does not require testing the controller in simulation or reality. In
general, for this optimization problem we can derive the exact
equation and find a solution with gradient-based optimization
\cite{fiacco1990nonlinear}. For the specific behavior space in the
example problem in this paper, though, the discretized search space of
the behavior-performance map is small enough that we can exhaustively compute the
acquisition value of each solution of the behavior-performance map and then choose
the maximum value.

Several different acquisition functions exist, such as the probability
of improvement, the expected improvement, or the Upper Confidence
Bound (UCB)\cite{brochu2010tutorial, calandra2014experimental}. We
chose UCB because it provided the best results in several previous
studies\cite{brochu2010tutorial, calandra2014experimental}. The
equation for UCB is:
\begin{equation}
\mathbf{x}_{t+1}= \operatorname*{arg\,max}_\mathbf{x} (\mu_{t}(\mathbf{x})+ \kappa\sigma_t(\mathbf{x}))
\label{ucb}
\end{equation}
where $\kappa$ is a user-defined parameter that tunes the tradeoff between exploration and exploitation.

The acquisition function handles the exploitation/exploration
trade-off of the adaptation (M-BOA) step. In the UCB function
(Eq. \ref{ucb}), the emphasis on exploitation vs. exploration is
explicit and easy to adjust. The UCB function can be seen as the
maximum value (argmax) across all solutions of the weighted sum of the
expected performance (mean of the Gaussian, $\mu_{t}(\mathbf{x})$) and
of the uncertainty (standard deviation of the Gaussian,
$\sigma_t(\mathbf{x})$) of each solution. This sum is weighted by the
$\kappa$ factor. With a low $\kappa$, the algorithm will choose
solutions that are expected to be high-performing. Conversely, with a
high $\kappa$, the algorithm will focus its search on unexplored areas
of the search space that may have high-performing solutions. The
$\kappa$ factor enables fine adjustments to the
exploitation/exploration trade-off of the M-BOA algorithm (the adaptation step). We describe how we chose the $\kappa$ value in supplementary methods, section \ref{sec:kappa}.

\paragraph{Code availability} The source code (for GNU/Linux) for the experiments of this paper is available at the following URL:\\
{\footnotesize
\url{http://pages.isir.upmc.fr/~mouret/code/ite_source_code.tar.gz}} An implementation of the Bayesian optimization algorithm is freely available on:\\\url{http://github.com/jbmouret/limbo}



\subsection{Hexapod Experiment}

\paragraph{Physical robot}\label{sec:robot}
The robot is a 6-legged robot with 3 degrees of freedom (DOFs) per
leg. Each DOF is actuated by position-controlled servos (MX-28
Dynamixel actuators manufactured by Robotis). The first servo controls
the horizontal (front-back) orientation of the leg and the two others
control its elevation. An RGB-D camera (Xtion, from ASUS) is fixed on
top of the robot. Its data are used to estimate the forward
displacement of the robot via an RGB-D SLAM
algorithm\footnote{\url{http://wiki.ros.org/ccny_openni_launch}}\cite{dryanovski2013fast}
from the robot operating system (ROS)
framework\footnote{\url{http://www.ros.org}} \cite{Ros2009}.

\paragraph{Simulator} \label{sec:self-model}
The simulator is a dynamic physics simulation of the undamaged
6-legged robot on flat ground (Fig. \ref{fig:matrix}).  
We weighted each segment of the leg and the body of the real robot, and we used
the same masses for the simulations. The simulator is based on the
Open Dynamics Engine (ODE, \url{http://www.ode.org}).

\paragraph{Parametrized controller}\label{sec:controller} 
The angular position of each DOF is governed by a periodic function
$\gamma$ parametrized by its amplitude $\alpha$, its phase $\phi$, and
its duty cycle $\tau$ (the duty cycle is the proportion of one period
in which the joint is in its higher position).  The function is
defined with a square signal of frequency 1Hz, with amplitude
$\alpha$, and duty cycle $\tau$. This signal is then smoothed via a
Gaussian filter in order to remove sharp transitions, and is then
shifted according to the phase $\phi$.

Angular positions are sent to the servos every 30 ms. In order to keep
the ``tibia'' of each leg vertical, the control signal of the third
servo is the opposite of the second one.  Consequently, angles sent
to the $i^{th}$ leg are:
\begin{itemize}
\item $\gamma(t,\ \alpha_{i_1},\ \phi_{i_1},\ \tau_{i_1})$ for DOF 1 
\item $\gamma(t,\ \alpha_{i_2},\ \phi_{i_2},\ \tau_{i_2})$ for DOF 2
\item $-\gamma(t,\ \alpha_{i_2},\ \phi_{i_2},\ \tau_{i_2})$ for DOF 3
\end{itemize}
This controller makes the robot equivalent to a 12 DOF system, even
though 18 motors are controlled.

There are $6$ parameters for each leg ($\alpha_{i_1}$, $\alpha_{i_2}$,
$\phi_{i_1}$, $\phi_{i_2}$, $\tau_{i_1}$, $\tau_{i_2}$), therefore
each controller is fully described by $36$ parameters. Each parameter
can have one of these possible values: {0, 0.05, 0.1, ...
0.95, 1}.  
Different values for these $36$ parameters can produce
numerous different gaits, from purely quadruped gaits to classic
tripod gaits.

This controller is designed to be simple enough to show the
performance of the algorithm in an intuitive setup. Nevertheless, the
algorithm will work with any type of controller, including
bio-inspired central pattern generators\cite{sproewitz2008learning}
and evolved neural
networks\cite{Yosinski2011,Clune2011,clune2009evolving,lee2013evolving}.

\paragraph{Reference controller}\label{sec:ref_controller}
Our reference controller is a classic tripod gait\cite{Siciliano2008, Wilson1966, saranli2001rhex, schmitz2001biologically, ding2010locomotion, steingrube2010self}. It involves two tripods: legs 1-4-5 and legs 2-3-6 (Fig.~\ref{fig:matrix}). This controller is designed to always keep the robot balanced on at least one of these tripods. The walking gait is achieved by lifting one tripod, while the other tripod pushes the robot forward (by shifting itself backward). The lifted tripod is then placed forward in order to repeat the cycle with the other tripod. This gait is static, fast, and similar to insect gaits\cite{Wilson1966,Delcomyn1971}.

Table S\ref{table:param_hexapode} shows the 36 parameters of the
reference controller. The amplitude orientation parameters
($\alpha_{i_1}$) are set to $1$ to produce the fastest possible gait,
while the amplitude elevation parameters ($\alpha_{i_2}$) are set to a
small value (0.25) to keep the gait stable. The phase elevation
parameters ($\phi_{i_2}$) define two tripods: 0.25 for legs
2-3-5; 
0.75 for legs 1-4-5. 
To achieve a cyclic motion
of the leg, the phase orientation values ($\phi_{i_1}$) are chosen by
subtracting $0.25$ to the phase elevation values ($\phi_{i_2}$), plus
a $0.5$ shift for legs 1-3-5, 
which are on the left side of the
robot.  All the duty cycle parameters ($\tau_i$) are set to $0.5$ so
that the motors spend the same proportion of time in their two limit
angles. The actual speed of the reference controller is not
important for the comparisons made in this paper: it is simply
intended as a reference and to show that the performance of classic,
hand-programmed gaits tend to fail when damage occurs.

 \ctable[star,
   cap = {},
   caption = {Parameters of the reference controller.},
   label   = {table:param_hexapode},
]{lccccccc}{}{
\hline
\multicolumn{2}{l}{Leg number} & 1 & 2 & 3 & 4 & 5 & 6\\
\hline
\multirow{3}*{First joint}& $\alpha_{i_1}$& 1.00 & 1.00 & 1.00 & 1.00 & 1.00 & 1.00\\
&$\phi_{i_1}$& 0.00 & 0.00 & 0.50 & 0.50 & 0.00 & 0.00\\ 
& $\tau_{i_1}$& 0.5 & 0.5 & 0.5 & 0.5 & 0.5 & 0.5\\
\hline
\multirow{3}*{Two last joints}& $\alpha_{i_2}$& 0.25 & 0.25 & 0.25 & 0.25 & 0.25 & 0.25\\
&$\phi_{i_2}$& 0.75 & 0.25 & 0.25 & 0.75 & 0.75 & 0.25\\ 
&$\tau_{i_2}$& 0.5 & 0.5 & 0.5 & 0.5 & 0.5 & 0.5\\
\hline}

\paragraph{Random variation of controller's parameters} 
Each parameter of the controller has a
$5\%$ chance of being changed to any value in the set of possible
values, with the new value chosen randomly from a uniform distribution
over the possible values.

\paragraph{Main Behavioral descriptor (duty factor)}\label{sec:descriptor}
The default behavioral descriptor is a 6-dimensional vector that corresponds
to the proportion of time that each leg is in contact with the
ground (also called duty factor).  When a controller is simulated, the
algorithm records at each time step (every 30 ms) whether each leg is
in contact with the ground (1: contact, 0: no contact). The result is
6 Boolean time series ($C_i$ for the $i^{th}$ leg).  The behavioral
descriptor is then computed with the average of each time series:

\begin{equation}
\mathbf{x}= \left[\begin{array}{c}
    \frac{\sum_{t}C_1(t)}{\textrm{numTimesteps}(C_1)}\\
    \vdots\\
    \frac{\sum_{t}C_6(t)}{\textrm{numTimesteps}(C_6)}\end{array}\right]
\end{equation}

During the generation of the behavior-performance map, the behaviors are
stored in the maps's cells by discretizing each dimension of the
behavioral descriptor space with these five values: \{0, 0.25, 0.5,
0.75, 1\}. During the adaptation phase, the behavioral descriptors are
used with their actual values and are thus not discretized.

\paragraph{Alternative Behavioral descriptor (orientation)}
The alternative behavioral descriptor tested on the physical robot (we investigated many other descriptors in simulation: Supplementary Experiment S5) characterizes changes in the angular position of the robot during walking, measured as the proportion of $15$ms intervals that each of the pitch, roll and yaw angles of the robot frame are positive (three dimensions) and negative (three additional dimensions):

\begin{equation}
\begin{aligned} \mathbf{x}&=\left[\begin{array}{l} 
\frac{1}{K}\sum_{k}U(\Theta^T(k) - 0.005 \pi)\\ 
\frac{1}{K}\sum_{k}U(-\Theta^T(k) -  0.005 \pi)\\ 
\frac{1}{K}\sum_{k}U(\Psi^T(k) -  0.005 \pi)\\ 
\frac{1}{K}\sum_{k}U(-\Psi^T(k) -  0.005 \pi)\\ 
\frac{1}{K}\sum_{k}U(\Phi^T(k) -  0.005 \pi)\\ 
\frac{1}{K}\sum_{k}U(-\Phi^T(k) -  0.005 \pi)
\end{array}\right] \end{aligned}
\end{equation}

where $\Theta^T(k)$, $\Psi^T(k)$ and $\Phi^T(k)$ denote the pitch, roll and yaw angles, respectively, of the robot torso (hence $T$) at the end of interval $k$, and $K$ denotes the number of $15$ms intervals during the 5 seconds of simulated movement (here, $K=5s/0.015s \approx 334$). The unit step function $U(\cdot)$ returns $1$ if its argument exceeds $0$, and returns $0$ otherwise. To discount for insignificant motion around $0$ rad, orientation angles are only defined as positive if they exceed $0.5\%$ of $\pi$ rad. Similarly, orientation angles are only defined as negative if they are less than $-0.5\%$ of $\pi$ rad.

\paragraph{Performance function}
\label{sec:measure}

In these experiments, the ``mission'' of the robot is to go forward as
fast as possible. The performance of a controller, which is a set of
parameters (section \ref{sec:controller}:
Parametrized controller),
 is defined as how far the
robot moves in a pre-specified direction in $5$ seconds.


During the behavior-performance map creation step, the performance is obtained
thanks to the simulation of the robot.  All odometry results reported
on the physical robot, during the adaptation step, are measured with
the embedded simultaneous location and mapping (SLAM)
algorithm\cite{dryanovski2013fast}. The accuracy of this algorithm was
evaluated by comparing its measurements to ones made by hand on 40
different walking gaits. These experiments revealed that the median
measurement produced by the odometry algorithm is reasonably accurate,
being just 2.2\% lower than the handmade measurement
(Extended Data Fig. \ref{fig:param_rho}d).

Some damage to the robot may make it flip over. In such cases, the visual
odometry algorithm returns pathological distance-traveled measurements
either several meters backward or forward. To remove these errors, we
set all distance-traveled measurements less than zero or greater than
two meters to zero. The result of this adjustment is that the
algorithm appropriately considers such behaviors
low-performing. Additionally, the SLAM algorithm sometimes reports
substantially inaccurate low values (outliers on Supplementary
Fig. \ref{fig:param_rho}d). In these cases the adaptation step algorithm
will assume that the behavior is low-performing and will select
another working behavior. Thus, the overall algorithm is not substantially
impacted by such infrequent under-measurements of performance.

\paragraph{Stopping criterion}\label{sec:stopping}
In addition to guiding the learning process to the most promising area
of the search space, the estimated performance of each solution in the
map also informs the algorithm of the maximum performance that
can be expected on the physical robot.  For example, if there is no
controller in the map that is expected to perform faster on the
real robot than 0.3m/s, it is unlikely that a faster solution
exists. This information is used in our algorithm to decide if it is
worth continuing to search for a better controller; if the algorithm
has already discovered a controller that performs nearly as well as
the highest value predicted by the model, we can stop the search.

Formally, our stopping criterion is
\begin{equation}
\max(\mathbf{P}_{1:t})\ge\alpha\max\limits_{\mathbf{x}\in \mathcal{P}}(\mu_t(\mathbf{x})),
\ \ \textrm{with }\alpha=0.9
\label{eq:stop}
\end{equation}

\noindent
where $\mathbf{x}$ is a location in the discrete behavioral space
(i.e. a type of behavior) and $\mu_t$ is the predicted performance of
this type of behavior.  Thus, when one of the tested solutions has a
performance of $90\%$ or higher of the maximum expected performance of
any behavior in the map, the algorithm terminates. At that
point, the highest-performing solution found so far will be the
compensatory behavior that the algorithm selects. An alternative way the
algorithm can halt is if 20 tests on the physical robot occur without
triggering the stopping criterion described in equation \ref{eq:stop}:
this event only occurred in 2 of 240 experiments performed on the
physical robot described in the main text. In this case, we selected
the highest-performing solution encountered during the search.
This user-defined stopping criterion is not strictly necessary, as the algorithm is guaranteed to stop in the worst case after every behavior in the map is tested, but it allows a practical limit on the number of trials performed on the physical robot.

\paragraph{Initiating the Adaptation Step}

The adaptation step is triggered when the performance drops by a certain amount. The simplest way to choose that threshold is to let the user specify it. Automating the selection of this value, and the impact of triggering the algorithm prematurely, is an interesting question for future research in this area.

\paragraph{Main parameters of MAP-Elites}
\begin{itemize}
\item parameters in controller: $36$
\item parameter values (controller): $0$ to $1$, with $0.05$ increments
\item size of behavioral space: $6$
\item possible behavioral descriptors: $\{0, 0.25, 0.5, 0.75, 1\}$
\item iterations: $40$ million
\end{itemize}

\paragraph{Main parameters of M-BOA}
\begin{itemize}
\item $\sigma_{noise}^2$: $0.001$
\item $\alpha$: $0.9$
\item $\rho$: $0.4$
\item $\kappa$: $0.05$
\end{itemize}

\subsection{Robotic Arm Experiment} 
\label{sec:arm_exp}
\paragraph{Physical robot}
The physical robot is a planar robotic arm with 8 degrees of freedom (Extended Data Fig. \ref{fig:arm}a) and a 1-degree-of-freedom gripper. The robot has to release a ball into a bin (a variant of the classic ``pick and place'' task in industrial robotics). To assess the position of the gripper, a red cap, placed on top of the gripper, is tracked with a video camera. The visual tracking is achieved with the ``cmvision'' ROS package, which tracks colored blobs (\url{http://wiki.ros.org/cmvision}). The eight joints of the robot are actuated by position-controlled servos manufactured by Dynamixel. To maximize the reliability of the the arm, the type of servo is not the same for all the joints: heavy-duty servos are used near the base of the robot and lighter ones are used for the end of the arm.  The first joint, fixed to the base, is moved by two MX-28 servos mounted in parallel. The second joint is moved by an MX-64 servo. The 3 subsequent servos are single MX-28s, and the 3 remaining servos are AX-18s. All the robot's joints are limited to a motion range of $\pm \pi / 2$.

\paragraph{Simulator} 
The generation of the behavior-performance map is made with a simulated robot in the same way as for the hexapod experiment. For consistency with the simulated hexapod experiments, we used the dynamic (as opposed to kinematic) version of the simulator, based on the ODE library. 
Any joint configuration that resulted in the arm colliding with itself was not added to the map. 

\paragraph{Parametrized controller}
The controller defines the target position for each joint. The controller is thus parametrized by eight continuous values from 0 to 1 describing the angle of each joint, which is mapped to the the total motion range of each joint of $\pm \pi / 2$. The 8 joints are activated simultaneously and are driven to their target position by internal PID controllers. 

We chose this simple control strategy to make the experiments easy to reproduce and highlight the contribution of Intelligent Trial \& Error for damage recovery. More advanced control strategies, for instance visual servoing\cite{Siciliano2008}, would be more realistic in a industrial environment, but they would have made it hard to analyze the experimental results because both Intelligent Trial \& Error and the controller would compensate for damage at the same time.

\paragraph{Randomly varying the controller's parameters} 
Each parameter of the controller (section ``Parametrized controller'') has a
$12.5\%$ chance of being changed to any value from 0 to 1, with the
new value chosen from a polynomial distribution as described on p. 124 of
(Deb, 2000), with $\eta_m=10.0$.

\paragraph{Behavioral descriptor}

Because the most important aspect of the robot's behavior in this task is the final position of the gripper, we use it as the behavioral descriptor:
\begin{equation}
	\mathrm{behavioral\_descriptor(simu(}\mathbf{c})) 
				= \left[
					\begin{array}{c} 
						x_g\\
						y_g
					\end{array}
				  \right]
\end{equation}
where $(x_g, y_g)$ denotes the position of the gripper once all the joint have reached their target position.

The size of the working area of the robot is a rectangle measuring $1.4m \times 0.7m$. For the behavior-performance map, this rectangle is discretized into a grid composed of $20000$ square cells ($200 \times 100$). The robot is $62$cm long.

\paragraph{Performance function}
Contrary to the hexapod experiment, for the robotic arm experiment the performance function for the behavior-map creation step and for the adaptation step are different. We did so to demonstrate that the two can be different, and to create a behavior-performance map that would work with arbitrary locations of the target bin.

For the \emph{behavior-performance map generation step} (accomplished via the MAP-Elites algorithm), the performance function captures the idea that all joints should contribute equally to the movement. Specifically, high-performance is defined as minimizing the variance of the joint angles, that is:

\begin{equation}
\mathrm{performance(simu}(\mathbf{c}))) = - \frac{1}{8} \sum_{i=0}^{i=7} \left(p_i - m\right)^2
\end{equation}
where $p_i$ is the angular position of joint $i$ (in radians) and $m = \frac{1}{8} \sum_{i=0}^{i=7} p_i$ is the mean of the joint angles.  This performance function does not depend on the target. The map is therefore generic: it contains a high-performing controller for each point of the robot's working space.

For the \emph{adaptation step} (accomplished via the M-BOA algorithm), the behavior-performance map, which is generic to many tasks, is used for a particular task. To do so, the adaption step has a different performance measure than the step that creates the behavior-performance map. For this problem, the predicted performance measure is the Euclidean distance to the target (closer is better). Specifically, for each behavior descriptor $\mathbf{x}$ in the map, performance is 

\begin{equation}
\mathcal{P}(\mathbf{x}) = - || \mathbf{x} - \mathbf{b} ||
\end{equation}
where $\mathbf{b}$ is the $(x,y)$ position of the target bin. Note that the variance of the joint angles, which is used to create the behavior-performance map, is ignored during the adaptation step.


The performance of a controller on the physical robot is minimizing the Euclidean distance between the gripper (as measured with the external camera) and the target bin:

\begin{equation}
\mathrm{performance(physical\_robot}(\mathcal{C}(\mathbf{\chi}))) = - ||\mathbf{x_g} - \mathbf{b}||
\end{equation}
where $\mathbf{x_g}$ is the position of the physical gripper after all joints have reached their final position, $\mathbf{b}$ is the position of the bin, and $\mathcal{C}(\mathbf{\chi})$ is the controller being evaluated ($\mathbf{\chi}$ is the position in simulation that controller reached).

If the gripper reaches a position outside of the working area, then the camera cannot see the marker. In these rare cases, we set the performance of the corresponding controller to  a low value ($-1$ m).

For the control experiments with traditional Bayesian optimization on the physical robot (see Supplementary Experiment S1), self-collisions are frequent during adaptation, especially given that we initialize the process with purely random controllers (i.e. random joint angles). While a single self-collision is unlikely to break the robot, hundreds of them can wear out the gearboxes because each servo continues to apply a force for a period of time until it determines that it cannot move. To minimize costs, and because we ran 210 independent runs of the algorithm (14 scenarios $\times$ 15 replicates), we first tested each behavior in simulation (taking the damage into account) to check that there were no self-collisions. If we detected a self-collision, the performance for that behavior was set to a low value ($-1$m). 

Auto-collisions are much less likely with Intelligent Trial \& Error because the behavior-performance map contains only controllers that do not self-collide on the undamaged, simulated robot. As a consequence, in the Intelligent Trial \& Error experiments we did not simulate controllers before testing them on the physical robot.

\paragraph{Stopping criterion}
Because the robot's task is to release a ball into a bin, the adaptation step can be stopped when the gripper is above the bin.  The
bin is circular with a diameter of 10 cm, so we stopped the adaptation step when the red cap is within 5 cm of the center of the
bin.

\paragraph{Main MAP-Elites parameters for the robotic arm experiment:}
\begin{itemize}
\item parameters in controller: $8$
\item controller parameter values: $0$ to $1$ (continuous)
\item dimensions in the behavioral space: $2$
\item simulated evaluations to create the behavior-performance map: $20$ million
\end{itemize}

\paragraph{Main M-BOA parameters for the robotic arm experiment:}
\begin{itemize}
\item $\sigma_{noise}^2$: $0.03$
\item $\rho$: $0.1$
\item $\kappa$: $0.3$
\end{itemize}

\begin{SI-figure*}
\centering
\includegraphics[width=0.9\linewidth]{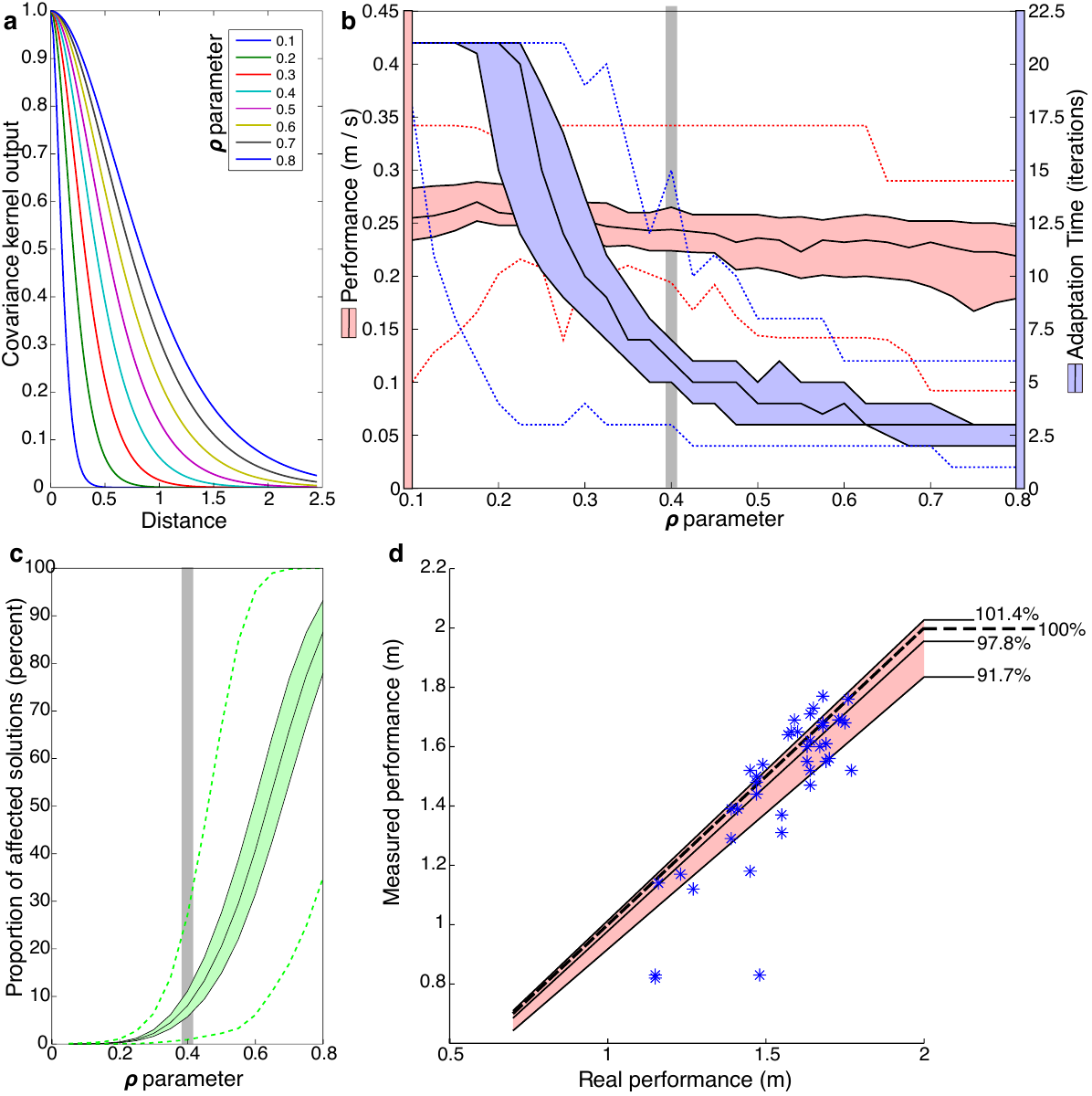}
\caption{\label{fig:param_rho}\textbf{The effect of changing the algorithm's parameters.} \textbf{(a) The shape of the Mat\'{e}rn kernel
  function for different values of the $\rho$ parameter.} \textbf{(b) Performance
  and required adaptation time obtained for different values of
  $\rho$.} For each $\rho$ value, the M-BOA algorithm was executed
  in simulation with 8 independently generated behavior-performance maps
  and for 6 different damage conditions (each case where one leg is missing).
 \textbf{(c) The number of controllers in the map affected by a new
  observation according to different values of the $\rho$
  parameter.} \textbf{(d) The precision of the odometry
  value}. The distances traveled by the physical robot, as measured manually (``real performance'') is compared to the measurements automatically provided by the
  simultaneous location and mapping (SLAM)   algorithm(``measured performance'').  The
  dashed black line indicates the hypothetical case where SLAM measurements are error-free and thus are the same as manual measurements. In (b), (c) and
  (d), the middle, black lines represent medians and the borders of
  the shaded areas show the 25\textsuperscript{th} and
  75\textsuperscript{th} percentiles. The dotted lines are the minimum
  and maximum values. The gray bars show the $\rho$ value chosen for
  the hexapod experiments in the main text. 
}
\end{SI-figure*}

\subsection{Selection of parameters}

All of the data reported in this section comes from experiments with the simulated hexapod robot, unless otherwise stated. 

\paragraph{Selecting the $\rho$ value}
\label{sec:rho}

For $\rho$ between 0.1 and 0.8, we counted the number of behaviors
from the map that would be influenced by a single test on the real hexapod
robot (we considered a behavior to be influenced when its predicted
performance was affected by more than $25\%$ of the magnitude of the
update for the tested behavior): with $\rho=0.2$, the update process does
not affect any neighbor in the map, with $\rho=0.4$, it affects
$10\%$ of the behaviors, and with $\rho=0.8$, it affects $80\%$ of
them. Additional values are shown in Extended Data Fig. \ref{fig:param_rho}c.

The previous paragraph describes tests we conducted to determine the number of behaviors in the map affected by different $\rho$ values, but those experiments do not tell us how different $\rho$ values affect the performance of the algorithm overall. To assess that, we then repeated the experiments from the main paper with a set of
possible values ($\rho\in[0.1:0.025:0.8]$) in simulation (i.e., with a
simulated, damaged robot), including testing on 6 separate damage
scenarios (each where the robot loses a different leg) with all 8
independently generated replicates of the default 6-dimensional behavior-performance map. The algorithm stopped if 20
adaptation iterations passed without success according to the stopping
criteria described in the main text and section \ref{sec:stopping}: Stopping criterion.
The results reveal that median performance decreases only modestly, but
significantly, when the value of $\rho$ increases: changing $\rho$
from 0.1 to 0.8 only decreases the median value $12\%$, from 0.25 m/s
to 0.22 m/s (p-value = $9.3 \times 10^{-5}$ via Matlab's Wilcoxon
ranksum test, Extended Data Fig. \ref{fig:param_rho}b). The variance in
performance, especially at the extreme low end of the distribution of
performance values, is not constant over the range of explored
values. Around $\rho=0.3$ the minimum performance
(Extended Data Fig. \ref{fig:param_rho}b, dotted red line) is higher than the
minimum performance for more extreme values of $\rho$.

A larger effect of changing $\rho$ is the amount of time required to
find a compensatory behavior, which decreases when the value of $\rho$
increases (Extended Data Fig. \ref{fig:param_rho}b). With a $\rho$ value lower than
0.25, the algorithm rarely converges in less than the allotted 20
iterations, which occurs because many more tests are required to cover
all the promising areas of the search space to know if a
higher-performing behavior exists than the best-already-tested. On the
other hand, with a high $\rho$ value, the algorithm updates its
predictions for the entire search space in a few observations: while
fast, this strategy risks missing promising areas of the search space.

In light of these data, we chose $\rho=0.4$ as the default value for
our hexapod experiments because it represents a good trade-off between
a high minimum performance and a low number of physical tests on the
robot. The value of $\rho$ for the robotic arm experiment has been chosen 
with the same method.

\paragraph{Selection of the $\kappa$ value}
\label{sec:kappa}
For the hexapod robot experiments, we chose $\kappa = 0.05$. This relatively low value
emphasizes exploitation over exploration. We chose this value because
the exploration of the search space has already been largely performed
during the behavior-performance map creation step: the map
suggests which areas of the space will be high-performing, and should
thus be tested, and which areas of the space are likely unprofitable,
and thus should be avoided.

For the robotic arm experiments, we chose $\kappa = 0.3$, which emphasizes exploration more, because it experimentally leads to better results.

\subsection{Running time}
\paragraph{Computing hardware}\label{sec:compHardware} 
All computation (on the physical robots and in simulation) was
conducted on a hyperthreaded 16-core computer (Intel Xeon E5-2650
2.00GHz with 64Gb of RAM). This computational power is mainly required
for the behavior-performance map creation step. Creating one map for the hexapod experiment took 2
weeks, taking advantage of the fact that map creation can
easily be parallelized across multiple cores. Map creation
only needs to be performed once per robot (or robot design), and can
happen before the robot is deployed. As such, the robot's onboard
computer does not need to be powerful enough to create the map.

For the hexapod robot experiment, the most expensive part of adaptation is the Simultaneous Localization And Mapping (SLAM) algorithm\cite{dryanovski2013fast, thrun2005probabilistic, dissanayake2001solution}, which measures the distance traveled on the physical hexapod robot. It is slow because it processes millions of 3D points per second. It can be run on less powerful computers, but doing so lowers its accuracy because fewer frames per second can be processed. As computers become faster, it should be possible to run high-accuracy SLAM algorithms in low-cost, onboard computers for robots.

The rest of the adaptation step needs much less computational
power and can easily be run on an onboard computer, such as a
smartphone. That is because it takes approximately 15,000 arithmetic
operations between two evaluations on the physical robot, which
requires less than a second or two on current smartphones.

\paragraph{Measuring how long adaptation takes (hexapod robot)}
The reported time to adapt includes the time required for the computer to select each test and the time to conduct each test on the physical robot. Overall, evaluating a controller on the physical hexapod robot takes about 8 seconds (median 8.03 seconds, 5\textsuperscript{th} and 95\textsuperscript{th} percentiles [7.95; 8.21] seconds): 0.5-1 second to initialize the robot, 5 seconds during which the robot can walk, 0.5-1 second to allow the robot to stabilize before taking the final measurement, and 1-2 seconds to run the SLAM algorithm. Identifying the first controller to test takes 0.03 [0.0216; 0.1277] seconds. The time to select the next controller to test increases depending on the number of previous experiments because the size of the Kernel Matrix (K matrix, see Methods and Extended Data Fig. \ref{fig:pseudo-code}), which is involved in many of the arithmetic operations, grows by one row and one column per test that has been conducted. For example, selecting the second test takes 0.15 [0.13; 0.22] seconds, while the $10^{th}$ selection takes 0.31 [0.17; 0.34] seconds.

\section{Supplementary Experiments S1}
\subsection*{Additional conditions for the robotic arm}

\paragraph{Methods}
We investigated 11 damage conditions on the physical robot in addition to the 3 described in the main text (Fig. 3). We used the same setup as described in the main text (see main text and section \ref{sec:arm_exp}). Extended Data Fig. \ref{fig:arm} shows the 14 scenarios.

For each of the 14 damage scenarios, we replicated experiments on the physical robot with 15 independently generated
behavior-performance maps (210 runs in total). We also replicated control experiments, which consist of traditional
Bayesian optimization directly in the original parameter space (i.e. without behavior-performance maps), 15 times for each of the 14 damage conditions (210 runs in total). For both the experimental and control treatments, each experiment involved 30 evaluations on the physical robot (31 if the first trial is counted). In many cases, not all 30 evaluations were required to reach the target, so we report only the number of trials required to reach that goal.

\paragraph{Results}

After running the MAP-Elites algorithm for 20 million evaluations, each of the 15 generated maps contain more than $11,000$ behaviors (11,209 [1,1206; 1,1217] behaviors, Extended Data Fig. \ref{fig:arm}c). 

In all the generated maps, the regions of different performance values for behaviors are arranged in concentric shapes resembling cardioids (inverted, heart-shaped curves) that cover the places the robot can reach (Extended Data Fig. \ref{fig:arm}c). The black line drawn over the shown map corresponds to all the positions of the end-effector for which all the degrees of freedom are set to the same angle (from $-\pi / 4$ to $+\pi/4$), that is, for the theoretically highest achievable performance (i.e. the lowest possible variance in servo angles). The performance of the behaviors tends to decrease the further they are from this optimal line.

The adaptation results (Extended Data Fig. \ref{fig:arm}e) show that the
Intelligent trial and error algorithm manages to reach the goal of being less than 5 cm
from the center of the bin for all the runs in all the tested scenarios save two (scenarios 11 \& 12). For these two scenarios, the algorithm still reaches the
target 60\% and 80\% of the time, respectively. For all the
damage conditions, the Intelligent Trial and Error algorithm reaches the target
significantly more often than the Bayesian optimization algorithm ($p<10^{-24}$).
Specifically, the median number of iterations to reach the target (Extended Data Fig. \ref{fig:arm}f) is below 11 iterations (27.5 seconds) for all scenarios except 11 and 12, for which 31 and 20 iterations are required, respectively. When the robot is not able to reach the target, the recorded number of iterations is set to 31, which explains why the median number of iterations for the Bayesian optimization algorithm is equal to 31 for most damage conditions. For all the damage conditions except one (scenario 11), the Intelligent Trial and Error algorithm used fewer trials to reach the target than the traditional Bayesian optimization algorithm.

If the robot is allowed to continue its experiment after reaching the
5 cm radius tolerance, for a total of 31 iterations
(Extended Data Fig. \ref{fig:arm}g), it reaches an accuracy around 1 cm for all
the damage conditions except the two difficult ones (scenarios 11 and
12). This level of accuracy is never achieved with the classic Bayesian
optimization algorithm, whose lowest median accuracy is $2.6cm$.

Scenarios 11 and 12 appear to challenge the Intelligent Trial and Error algorithm. 
While in both cases the success rate is improved, though not substantially, in case 11 the median accuracy is actually lower. 
These results stem from the fact that the difference between the successful pre-damage and post-damage behaviors is so large that the post-damage solutions for both scenarios lie outside of the map. This illustrates a limit of the proposed approach: if the map does not contain a behavior able to cope with the damage, the robot will not be able to adapt. This limit mainly comes from the behavioral descriptor choice: we chose it because of its simplicity, but it does not capture all of the important dimensions of variation of the robot. More sophisticated descriptors are likely to allow the algorithm to cope with such situations.  On the other hand, this experiment shows that with a very simple behavioral descriptor, using only the final position of the end-effector, our approach is able to deal with a large variety of different target positions and is significantly faster than the traditional Bayesian optimization approach (Extended Data Fig. \ref{fig:arm}d, maximum p-value over each time step $< 10^{-16}$), which is the current state of the art technique for direct policy search in robotics\cite{lizotte2007automatic, Tesch2011,calandra2014experimental, Kober2013}.

\section{Supplementary Experiments S2}
\subsection*{The contribution of each subcomponent of the Intelligent Trial and Error Algorithm}

\paragraph{Methods}
The Intelligent Trial and Error Algorithm relies on three main
concepts: (1) the creation of a behavior-performance map in simulation
via the MAP-Elites algorithm, (2) searching this map with a
Bayesian optimization algorithm to find behaviors that perform well on
the physical robot, and (3) initializing this Bayesian optimization
search with the performance predictions obtained via the MAP-Elites
algorithm: note that the second step could be performed without the
third step by searching through the MAP-Elites-generated behavior-performance
map with Bayesian optimization, but having the initial priors
uniformly set to the same value. We investigated the contribution of
each of these subcomponents by testing five variants of our algorithm
: in each of them, we deactivated one of
these three subcomponents or replaced it with an alternative algorithm
from the literature. We then tested these variants on the hexapod robot. The variants are as follows:
\begin{itemize} 
\item Variant 1 (MAP-Elites in 6 dimensions + random search):
  evaluates the benefit of searching the map via Bayesian
  optimization by searching that map with random search
  instead. Each iteration, a behavior is randomly selected from the
  map and tested on the robot. The best one is kept.
\item Variant 2 (MAP-Elites in 6 dimensions + Bayesian optimization,
  no use of priors): evaluates the contribution of initializing the
  Gaussian process with the performance predictions of the behavior-performance
  map. In this variant, the Gaussian process is initialized with a
  constant mean (the average performance of the map: $0.24$ m/s) at
  each location in the behavior space and a constant variance (the
  average variance of the map's performance:
  $0.005$~$m^2/s^2$). As is customary, the first few trials (here, 5) of the Bayesian optimization process are selected randomly instead of letting the algorithm choose those points, which is known to improve performance.\cite{calandra2014experimental}
\item Variant 3 (MAP-Elites in 6 dimensions + policy gradient):
  evaluates the benefit of Bayesian optimization compared to a more
  classic, local search algorithm\cite{Kober2013,kohl2004policy};
  there is no obvious way to use priors in policy gradient algorithms.
\item Variant 4 (Bayesian optimization in the original parameter space
  of 36 dimensions): evaluates the contribution of using a map
  in a lower-dimensional behavioral space. This variant searches
  directly in the original 36-dimensional parameter space instead of
  reducing that space to the lower-dimensional (six-dimensional)
  behavior space. Thus, in this variant no map of behaviors is
  produced ahead of time: the algorithm searches directly in the
  original, high-dimensional space. This variant corresponds to one of
  the best algorithms known to learn locomotion
  patterns\cite{lizotte2007automatic,calandra2014experimental}. In
  this variant, the Gaussian process is initialized with a constant
  mean set to zero and with a constant variance ($0.002 m^2/s^2$). As described above, the five first trials are selected from pure random search to prime the
    Bayesian optimization algorithm\cite{calandra2014experimental}.
\item Variant 5 (Policy gradient in the original parameter space of 36
  dimensions): a stochastic gradient descent in the original parameter
  space\cite{kohl2004policy}. This approach is a classic reinforcement
  learning algorithm for locomotion\cite{Kober2013} and it is a
  baseline in many papers\cite{lizotte2007automatic}.
\end{itemize}

It was necessary to compare these variants in simulation because doing
so on the physical robot would have required months of experiments and
would have repeatedly worn out or broken the robot. We modified the
simulator from the main experiments (section \ref{sec:self-model}: Simulator) to
emulate 6 different possible damage conditions, each of which involved
removing a different leg. For variants in which MAP-Elites creates a
map (variants 1, 2 and 3), we used the same maps from
the main experiments (the eight independently generated maps,
which were all generated with a simulation of the undamaged robot): In
these cases, we launched ten replicates of each variant for each of
the eight maps and each of the six damage conditions. There are
therefore $10\times8\times6=480$ replicates for each of those
variants. For the other variants (4 and 5), we replicated each
experiment 80 times for each of the six damage conditions, which also
led to $80\times6=480$ replicates per variant. In all these simulated
experiments, to roughly simulate the distribution of noisy odometry
measurements on the real robot, the simulated performance values were
randomly perturbed with a multiplicative Gaussian noise centered on
0.95 with a standard deviation of 0.1.

We analyze the fastest walking speed achieved with each variant after
two different numbers of trials: the first case is after 17 trials,
which was the maximum number of iterations used by the Intelligent
Trial and Error Algorithm, and the second case is after 150 trials, which is
approximately the number of trials used in previous work\cite{kohl2004policy,lizotte2007automatic,calandra2014experimental}.

\paragraph{Results}
After 17 trials on the robot, Intelligent Trial and Error
significantly outperforms all the variants
(Extended Data Fig. \ref{fig:control_exp}b, $p < 10^{-67}$, Intelligent
Trial and Error performance: 0.26 [0.20; 0.33] m/s), demonstrating
that the three main components of the algorithm are needed to quickly
find high-performing behaviors. Among the investigated variants, the
random search in the map performs the best (Variant 1: 0.21
[0.16; 0.27] m/s), followed by Bayesian optimization in the map
(Variant 2: 0.20 [0.13; 0.25] m/s), and policy gradient in the
map (Variant 3: 0.13 [0; 0.23] m/s). Variants that search
directly in the parameter space did not find any working behavior
(Variant 4, Bayesian optimization: 0.04m/s, [0.01; 0.09]; Variant 5,
policy gradient: 0.02 [0; 0.06] m/s).

There are two reasons that random search performs better than one
might expect. First, the map only contains high-performing
solutions, which are the result of the intense search of the
MAP-Elites algorithm (40 million evaluations in simulation). The
map thus already contains high-performing gaits of nearly every
possible type. Therefore, this variant is not testing random
controllers, but is randomly selecting high-performing
solutions. Second, Bayesian optimization and policy gradient are not
designed for such a low number of trials: without the priors on
performance predictions introduced in the Intelligent Trial and Error
Algorithm, the Bayesian optimization process needs to learn the
overall shape of the search space to model it with a Gaussian
process. 17 trials is too low a number to effectively sample six
dimensions (for a uniform sampling with only two possible values in
each dimension, $2^6=64$ trials are needed; for five possible values,
$5^6=15,625$ samples are needed). As a consequence, with this low
number of trials, the Gaussian process that models the performance
function is not informed enough to effectively guide the search. For
the policy gradient algorithm, a gradient is estimated by empirically
measuring the partial derivative of the performance function in each
dimension. To do so, following\cite{kohl2004policy}, the policy
gradient algorithm performs 15 trials at each iteration. Consequently,
when only 17 trials are allowed, it iterates only once. In addition,
policy gradient is a local optimization algorithm that highly depends
on the starting point (which is here chosen randomly), as illustrated
by the high variability in the performance achieved with this variant
(Extended Data Fig. \ref{fig:control_exp}b).

The issues faced by Bayesian optimization and policy gradient are
exacerbated when the algorithms search directly in the original,
36-dimensional parameter space instead of the lower-dimensional
(six-dimensional) behavior space of the map. As mentioned
previously, no working controller was found in the two variants
directly searching in this high-dimensional space.

Overall, the analysis after 17 trials shows that:
\begin{itemize}
\item The most critical component of the Intelligent Trial and Error
  Algorithm is the MAP-Elites algorithm, which reduces the search
  space and produces a map of high-performing behaviors in that
  space: $p< 5 \times 10^{-50}$ when comparing variants searching in
  the behavior-performance map space vs. variants that search in the
  original, higher-dimensional space of motor parameters.
\item Bayesian optimization critically improves the search, but only
  when it is initialized with the performance obtained in simulation
  during the behavior-performance map creation step (with initialization:
  0.26 [0.20; 0.33] m/s, without initialization: 0.20 [0.13; 0.25]
  m/s, $p=10^{-96}$).
\end{itemize}
 To check whether these variants might perform better if allowed the number of evaluations typically given to previous state-of-the-art algorithms\cite{kohl2004policy,lizotte2007automatic,calandra2014experimental}, we continued the experiments until 150 trials on the robot were conducted (Extended Data Fig. \ref{fig:control_exp}c). Although the results for all the variants improved, Intelligent Trial and Error still outperforms all them ($p < 10^{-94}$; Intelligent Trial and Error: 0.31 [0.26; 0.37] m/s, random search: 0.26 [0.22; 0.30] m/s, Bayesian optimization: 0.25 [0.18; 0.31] m/s, policy search: 0.23 [0.19, 0.29] m/s). These results are consistent with the previously published results\cite{kohl2004policy,lizotte2007automatic,calandra2014experimental,Kober2013}, which optimize in 4 to 10 dimensions in a few hundred trials. Nevertheless, when MAP-Elites is not used, i.e. when we run these algorithms in the original 36 dimensions for 150 evaluations, Bayesian optimization and policy gradient both perform much worse (Bayesian optimization: 0.08 [0.05; 0.12]; policy gradient: 0.06 [0.01; 0.12] m/s). These results shows that MAP-Elites is a powerful method to reduce the dimensionality of a search space for learning algorithms, in addition to providing helpful priors about the search space that speed up Bayesian optimization.

 Overall, these additional experiments demonstrate that each of the three main components of the Intelligent Trial and Error Algorithm substantially improves performance. The results also indicate that Intelligent Trial and Error significantly outperforms previous algorithms for both damage recovery
\cite{erden2008free,bongard2006resilient,christensen2013fault,mahdavi2006innately,koos2013fast} and gait learning \cite{hornby2005autonomous,kohl2004policy,barfoot2006experiments,sproewitz2008learning,koos2013transferability,lizotte2007automatic,Tesch2011,calandra2014experimental,Kober2013,Yosinski2011}, and can therefore be considered the state of the art.

\section{Supplementary Experiments S3}
\raggedbottom %
\subsection*{Robustness to environmental changes}

\paragraph{Methods}
The map creation algorithm (MAP-Elites) uses an undamaged robot
on flat terrain. The main experiments show that this algorithm
provides useful priors for damage recovery on a flat terrain. In these
supplementary experiments, we evaluated, in simulation, if the
map created on flat terrain also provides a useful starting
point for discovering gaits for sloped terrains.

We first evaluated the effect slopes have on undamaged robots
(Extended Data Fig. \ref{fig:exp_slope}a). We launched 10 replicates for each of
the eight maps and each one-degree increment between
$-20^{\circ}$ and $+20^{\circ}$, for a total of $10\times 8 \times
41=3280$ experiments. As in Supplementary Experiments S2, to roughly
simulate the distribution of noisy odometry measurements on the real
robot, we perturbed performance values with a multiplicative Gaussian
noise centered on 0.95 with a standard deviation of 0.1.

\paragraph{Results}
The results show that, when the slope is negative (descending), the
Intelligent Trial and Error approach finds fast gaits in fewer than
than 3 trials. For reference, a hand-designed, classic, tripod gait
(section \ref{sec:controller}) falls on slopes below $-15^{\circ}$
degrees. When the slope is positive (ascent), Intelligent Trial and
Error finds slower behaviors, as is expected, but even above
$10^{\circ}$ the gait learned by Intelligent Trial and Error
outperforms the reference gait on flat ground. Overall, for every
slope angle, the controller found by Intelligent Trial and Error is
faster than the hand-designed reference controller.

We further evaluated damage recovery performance for these same
slopes with the same setup as Experiments S2 (6 damage
conditions). We launched 10 replicates for each damage condition, for
8 independently generated behavior-performance maps, and each two-degree increment between $-20^{\circ}$ and
$+20^{\circ}$ degrees. There are therefore 480 replicates for each
two-degree increment between $-20^{\circ}$ and $+20^{\circ}$, for a total of $480
\times 21 = 10080$ experiments.

Intelligent Trial and Error is not critically
affected by variations of slope between $-10^{\circ}$ and
$+10^{\circ}$ (Extended Data Fig. \ref{fig:exp_slope}b): for these slopes, and for
all 6 damage conditions, Intelligent Trial and Error finds fast gaits
(above 0.2 m/s) in less than 15 tests on the robot despite
the slope. As expected, it finds faster gaits for negative slopes
(descent) and slower gaits for positive slopes (ascent). For slopes
below $-10^{\circ}$ and above $10^{\circ}$, the algorithm performs
worse and requires more trials. These results likely are caused by the
constraints placed on the controller and the limited sensors on the
robot, rather than the inabilities of the algorithm. Specifically, the
controller was kept simple to make the science clearer, more
intuitive, and more reproducible. Those constraints, of course,
prevent it from performing the more complex behaviors necessary to
deal with highly sloped terrain. For example, the constraints prevent
the robot from keeping its legs vertical on sloped ground, which would
substantially reduce slippage.  Nevertheless, the median Intelligent
Trial and Error compensatory gait still outperforms the median
performance of the reference gait on all slope angles.

\section{Supplementary Experiments S4}
\subsection*{Comparison between MAP-Elites and Random Sampling}
\paragraph{Methods}
The MAP-Elites algorithm is a stochastic search algorithm that attempts to fill a discretized map with the highest-performing solution at each point in the map. As explained in the main text, each point in the map represents a different type of behavior, as defined by the behavioral dimension of the map. MAP-Elites generates new candidate points by randomly selecting a location in the map, changing the parameters of the controller that is stored there, and then saving that controller in the appropriate map location if it is better than the current occupant at that location. Intuitively, generating new candidate solutions from the best solutions found so far should be better than generating a multitude of controllers randomly and then keeping the best one found for each location in the map. In this section we report on experiments that confirm that intuition. 

To understand the advantages of MAP-Elites over random sampling, we compared the two algorithms by generating data with the simulated hexapod. The experiments have the same virtual robot, environment,  controller, performance function, and behavioral descriptors as in the main experiments (see Methods). We analyzed the number of cells for which a solution is found (an indication of the diversity of behavior types the algorithms generate), the average performance of behaviors in the map, and the maximum performance discovered.

We replicated each experiment 8 times, each of which included 20 million evaluations on the simulated robot.




\begin{SI-figure*}
\centering
\includegraphics[width=0.8\linewidth]{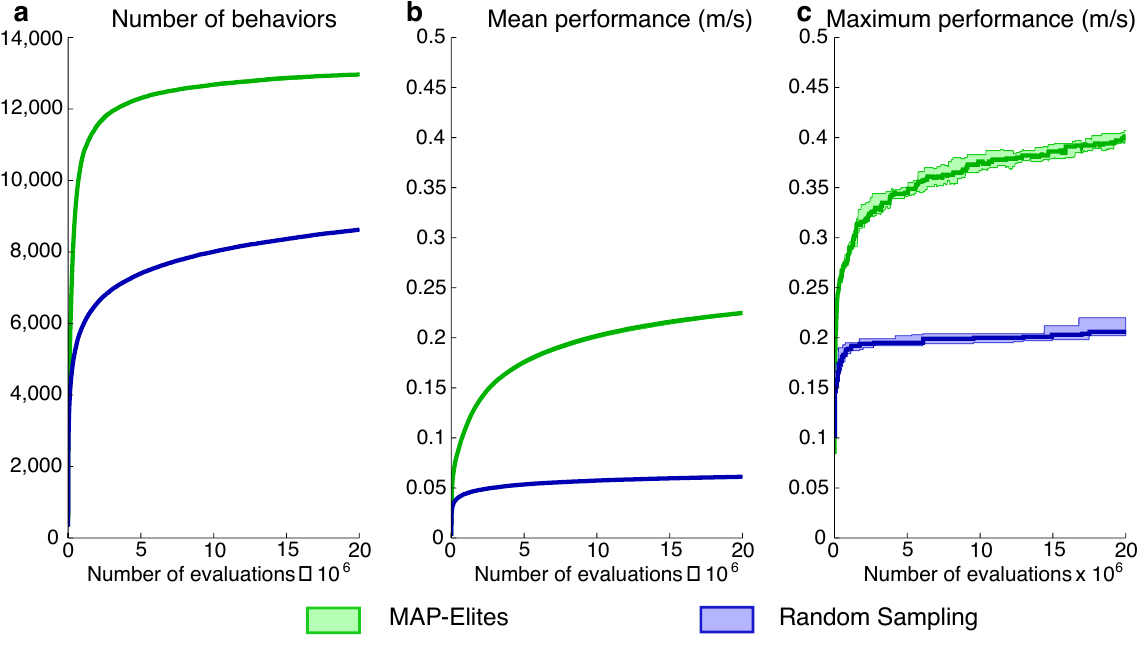}
\caption{\label{fig:map-elites} \textbf{Comparing MAP-Elites and
    random sampling for generating behavior-performance maps.} \textbf{(a) The number of points in the map for
    which a behavior is discovered.} \textbf{(b) The mean performance
    of the behaviors in the map.} \textbf{(c) The maximum performance
    of the behaviors in the map.} For all these figures, the middle
  lines represent medians over 8 independently generated maps and the shaded regions extend to the
  25\textsuperscript{th} and 75\textsuperscript{th} percentiles,
  even for (a) and (b), where the variance of the distribution is so
  small that it is difficult to see. See Supplementary Experiment S4
  for methods and analysis.} 
\end{SI-figure*}

\paragraph{Results}
The results show that the MAP-Elites algorithm outperforms random sampling on each of these measures (Extended Data Fig. \ref{fig:map-elites}). After 20 million evaluations, about 13000 cells (median: 12968,  $5^{th}$ \& $95^{th}$ percentiles: [12892; 13018]) are filled by MAP-Elites (about 83\% percent of the map), whereas random sampling only filled approximately 8600 (8624 [8566; 8641]) cells  (about 55\% percent of the map) (Extended Data Fig. \ref{fig:map-elites}a). The difference between the two algorithms is large and appears early (Extended Data Fig. \ref{fig:map-elites}a); after only 1 million evaluations, MAP-Elites filled 10670 [10511; 10775] cells (68\% of the map), whereas random sampling filled 5928 [5882; 5966] cells (38\% of the map). 

The solutions discovered by MAP-Elites are not only more numerous, but also outperform those found by random sampling (Extended Data Fig. \ref{fig:map-elites}b): with MAP-Elites, after 20 million evaluations the average performance of filled cells is 0.22 [0.22; 0.23] m/s, whereas it is 0.06 [0.06; 0.06] m/s with random sampling, which is similar to the performance obtained with the reference controller on a damaged robot (Fig. 3). These two results demonstrate that MAP-Elites is a much better algorithm than random sampling to find a map of the diverse, ``elite'' performers in a search space. 

In addition, MAP-Elites is a better optimization algorithm, as measured by the performance of the best single solution produced. The performance of the best solution in the map after 20 million evaluations is 0.40 [0.39;0.41] m/s with MAP-Elites, compared to 0.21 [0.20; 0.22] m/s with random sampling.


\section{Supplementary Experiments S5}

\label{sec:alt_bd}
\subsection*{Alternative behavioral descriptors}

\paragraph{Methods}

To create a map with MAP-Elites, one has to define the dimensions of the behavioral space, i.e. the behavioral descriptors. The main experiments show that using a predefined behavioral descriptor (the proportion of time that each leg of a hexapod robot is in contact with the ground, i.e. the duty factor) creates a map that provides useful priors for damage recovery. 

This section describes how we tested (in simulation) how performance is affected by alternative behavioral descriptors, including descriptors that have a different number of dimensions. We also evaluated how performance is affected if the behavioral descriptors are randomly selected from a large list of potential descriptors. This test simulates the algorithm's performance if the behavioral descriptors are chosen without insight into the problem domain. 

The behavioral descriptors we tested are as follows:
\begin{enumerate} 
\item Duty factor (6-dimensional): This descriptor is the default one from the main experiment. It corresponds to the proportion of time each leg is in contact with the ground:

\begin{equation}
\begin{aligned} \mathbf{x}&= \left[\begin{array}{c} \frac{\sum_{t}C_1(t)}{\textrm{numTimesteps}}\\  \vdots\\    \frac{\sum_{t}C_6(t)}{\textrm{numTimesteps}}\end{array}\right] 
\end{aligned}
\end{equation}

where $C_i(t)$ denotes the Boolean value of whether leg $i$ is in contact with the ground at time $t$ (1: contact, 0: no contact).

\item Orientation (6-dimensional): {This behavioral descriptor characterizes changes in the angular position of the robot during walking, measured as the proportion of $15$ms intervals that each of the pitch, roll and yaw angles of the robot frame are positive (three dimensions) and negative (three additional dimensions):}

\begin{equation}
\begin{aligned} \mathbf{x}&=\left[\begin{array}{l} 
\frac{1}{K}\sum_{k}U(\Theta^T(k) - 0.005 \pi)\\ 
\frac{1}{K}\sum_{k}U(-\Theta^T(k) -  0.005 \pi)\\ 
\frac{1}{K}\sum_{k}U(\Psi^T(k) -  0.005 \pi)\\ 
\frac{1}{K}\sum_{k}U(-\Psi^T(k) -  0.005 \pi)\\ 
\frac{1}{K}\sum_{k}U(\Phi^T(k) -  0.005 \pi)\\ 
\frac{1}{K}\sum_{k}U(-\Phi^T(k) -  0.005 \pi)
\end{array}\right] \end{aligned}
\end{equation}

where $\Theta^T(k)$, $\Psi^T(k)$ and $\Phi^T(k)$ denote the pitch, roll and yaw angles, respectively, of the robot torso (hence $T$) at the end of interval $k$, and $K$ denotes the number of $15$ms intervals during the 5 seconds of simulated movement (here, $K=5s/0.015s \approx 334$). The unit step function $U(\cdot)$ returns $1$ if its argument exceeds $0$, and returns $0$ otherwise. To discount for insignificant motion around $0$ rad, orientation angles are only defined as positive if they exceed $0.5\%$ of $\pi$ rad. Similarly, orientation angles are only defined as negative if they are less than $-0.5\%$ of $\pi$ rad.

\item Displacement (6-dimensional): This behavioral descriptor characterizes changes in the postion of the robot during walking. It is measured as the proportion of $15$ms intervals that the robot is positively or negatively displaced along each of the $x$, $y$, and $z$ axes: 

\begin{equation}
\begin{aligned} \mathbf{x}&=\left[\begin{array}{l} 
\frac{1}{K}\sum_{k}U(\Delta x(k) - 0.001)\\ 
\frac{1}{K}\sum_{k}U(-\Delta x(k) - 0.001)\\ 
\frac{1}{K}\sum_{k}U(\Delta y(k) - 0.001)\\ 
\frac{1}{K}\sum_{k}U(-\Delta y(k) - 0.001)\\ 
\frac{1}{K}\sum_{k}U(\Delta z(k) - 0.001)\\ 
\frac{1}{K}\sum_{k}U(-\Delta z(k) - 0.001)\end{array}\right] \end{aligned}
\end{equation}

where $\left[\Delta x(k), \Delta y(k), \Delta z(k)\right]$ denote the linear displacement in meters of the robot during interval $k$, and $K$ denotes the number of $15$ms intervals during 5 seconds of simulated movement (here, $K=5s/0.015s \approx 334$). The unit step function $U(\cdot)$ returns a value of $1$ if its argument exceeds $0$, and returns a value of $0$ otherwise. To ignore insignificant motion, linear displacements are defined to be positive if they exceed $1$mm, and are defined to be negative if they are less than $-1$mm.

\item Total energy expended per leg (6-dimensional): This behavioral descriptor captures the total amount of energy expended to move each leg during $5$ seconds of movement: 

\begin{equation}
\begin{aligned} \mathbf{x}&=\left[\begin{array}{c} 
\frac{E_1}{M_E}\\ 
\vdots\\ 
\frac{E_6}{M_E}
\end{array}\right] \end{aligned}
\end{equation}

where $E_i$ denotes the energy utilized by leg $i$ of the robot during 5 seconds of simulated movement, measured in N.m.rad. $M_E$ is the maximum amount of energy available according to the servo model of the simulator, which for 5 seconds is 100 N.m.rad.

\item Relative energy expended per leg (6-dimensional): This behavioral descriptor captures the amount of energy expended to move each leg relative to the energy expended by all the legs during $5$ seconds of simulated movement:

\begin{equation}
\begin{aligned} \mathbf{x}&=\left[\begin{array}{c} 
\frac{E_1}{\sum_{i=1..6}E_i}\\ 
\vdots\\ 
\frac{E_6}{\sum_{i=1..6}E_i}
\end{array}\right] \end{aligned}
\end{equation}

where $E_i$ denotes the energy utilized by leg $i$ of the robot during 5 seconds of simulated movement, measured in N.m.rad.

\item Deviation (3-dimensional): This descriptor captures the range of deviation of the center of the robot frame versus the expected location of the robot if it traveled in a straight line at a constant speed.

\begin{equation}
\label{eq:bd_amplitude}
\begin{aligned} \mathbf{x}&=\left[\begin{array}{c} 
\frac{0.95\left(\underset{t}\max(x(t)) - \underset{t}\min(x(t))\right)}{0.2}\\
\frac{0.95\left(\underset{t}\max(y(t) - \frac{y_\textrm{final}}{5}\times t) - \underset{t}\min(y(t) - \frac{y_\textrm{final}}{5}\times t)\right)}{0.2}\\
\frac{0.95\left(\underset{t}\max(z(t)) - \underset{t}\min(z(t))\right)}{0.2}
\end{array}\right] \end{aligned} 
\end{equation}

where $\left[x(t), y(t), z(t)\right]$ denote the position of robot's center at time $t$, and $\left[x_\textrm{final}, y_\textrm{final}, z_\textrm{final}\right]$ denote its final position after 5 seconds.

The robot's task is to move along the y-axis. Its starting position is (0,0,0). The deviation along the $x$ and $z$ axes is computed as the maximum difference in the robot's position in those dimensions at any point during 5 seconds. For the $y$ axis, $\frac{y_\textrm{final}}{5}$ corresponds to the average speed of the robot (the distance covered divided by total time), therefore $\frac{y_\textrm{final}}{5}\times t$ is the expected position at timestep $t$ if the robot was moving at constant speed. The deviation from the $y$ axis is computed with respect to this ``theoretical'' position.

To obtain values in the range [0,1], the final behavioral descriptors are multiplied by $0.95$ and then divided by $20$ cm (these values were determined empirically).

\item {Total ground reaction force per leg} (6-dimensional): This behavioral descriptor corresponds to the amount of force each leg applies to the ground, measured as a fraction the total possible amount of force that a leg could apply to the ground. Specifically, the measurement is

\begin{equation}
\begin{aligned} \mathbf{x}&=\left[\begin{array}{c} 
\frac{F_1}{M_F}\\ 
\vdots\\ 
\frac{F_6}{M_F}
\end{array}\right] \end{aligned}
\end{equation}

where $F_i$ denotes the ground reaction force (GRF) each leg $i$ generates, averaged over 5 seconds of simulated movement. $M_F$ is the maximum such force that each leg can apply, which is $10$N.

\item {Relative ground reaction force per leg} (6-dimensional): This behavioral descriptor corresponds to the amount of force each leg applies to the ground, relative to that of all the legs:

\begin{equation}
\begin{aligned} \mathbf{x}&=\left[\begin{array}{c} 
\frac{F_1}{\sum_{i=1..6}F_i}\\ 
\vdots\\ 
\frac{F_6}{\sum_{i=1..6}F_i}
\end{array}\right] \end{aligned}
\end{equation}

where $F_i$ denotes the ground reaction force (GRF) each leg $i$ generates, averaged over 5 seconds of simulated movement.

\item Lower-leg pitch angle (6-dimensional): This descriptor captures the pitch angle for the lower-leg with respect to the ground (in a global coordinate frame), averaged over 5 seconds:

\begin{equation}
\begin{aligned} \mathbf{x}&=\left[\begin{array}{c} 
\frac{\sum_{t}\Theta^L_1(t)}{\pi \times N_1}\\  
\vdots\\    
\frac{\sum_{t}\Theta^L_6(t)}{\pi \times N_6}
\end{array}\right] \end{aligned}
\end{equation}

where $\Theta^L_i(t)$ is the pitch angle of lower-leg $i$ (hence the $L$ in $\Theta^L_i$) when it is in contact with the ground at time $t$, and $N_i$ is the number of time-steps for which lower-leg $i$ touches the ground. The foot pitch angles are in range $\left[0, \pi\right]$ (as the leg can not penetrate the ground) and normalized to $\left[0, 1\right]$. 

\item Lower-leg roll angle (6-dimensional): This descriptor captures the roll angle for the lower-leg with respect to the ground (in a global coordinate frame), averaged over 5 seconds:

\begin{equation}
\begin{aligned}  \mathbf{x}&=\left[\begin{array}{c} 
\frac{\sum_{t}\Psi^L_1(t)}{\pi \times N_1}\\  
\vdots\\    
\frac{\sum_{t}\Psi^L_6(t)}{\pi \times N_6}
\end{array}\right] \end{aligned}
\end{equation}

where $\Psi^L_i(t)$ is the roll angle of lower-leg $i$ (hence $L$ in $\Psi^L_i$) when it is in contact with the ground at time $t$, and $N_i$ is the number of time-steps for which lower-leg $i$ touches the ground. The foot roll angles are in range $\left[0, \pi\right]$ (as the leg can not penetrate the ground) and normalized to $\left[0, 1\right]$. 

\item Lower-leg yaw angle (6-dimensional): This descriptor captures the yaw angle for the lower-leg with respect to the ground (in a global coordinate frame), averaged over 5 seconds:

\begin{equation}
\begin{aligned} \mathbf{x}&=\left[\begin{array}{c} 
\frac{\sum_{t}\Phi^L_1(t) + \pi}{2 \pi \times N_1}\\  
\vdots\\    
\frac{\sum_{t}\Phi^L_6(t) + \pi}{2 \pi \times N_6}
\end{array}\right] \end{aligned}
\end{equation}

where $\Phi^L_i(t)$ is the yaw angle of lower-leg $i$ (hence $L$ in $\Phi^L_i$) when it is in contact with the ground at time $t$,and $N_i$ is the number of time-steps for which lower-leg $i$ touches the ground. The foot yaw angles are in range $\left[-\pi, \pi\right]$ and are normalized to $\left[0, 1\right]$. 

\item Random (6-dimensional): The random behavioral descriptor differs from the other intentionally chosen descriptors in that it does not consist of one type of knowledge, but is instead randomly selected as a subset of variables from the previously described $11$ behavioral descriptors. This descriptor is intended to simulate a situation in which one has little expectation for which behavioral descriptor will perform well, so one quickly picks a few different descriptor dimensions without consideration or experimentation. Instead of generating one such list in this fashion, we randomly sample from a large set to find the average performance of this approach over many different possible choices. 

For the random descriptor, each of the 6-dimensions is selected at random (without replacement) from the $1\times3+10\times6=63$ available behavior descriptor dimensions described in the previous 11 descriptors (1 of the above descriptors is three-dimensional and the other 10 are six-dimensional): 

\begin{equation}
\begin{aligned} \mathbf{x}&=\left[\begin{array}{c} 
R_1\\ 
\vdots\\ 
R_6\end{array}\right] 
\end{aligned}
\end{equation}

where $R_i$ denotes the $i$\textsuperscript{th} dimension of the descriptor, randomly selected uniformly and without replacement from the $63$ available dimensions in behavior descriptors 1-11. 
\end{enumerate} 

It was necessary to compare these behavioral descriptors in simulation because doing so on the physical robot would have required months of experiments and would have repeatedly worn out or broken the robot. We modified the simulator from the main experiments (section \ref{sec:self-model}) to emulate 6 different possible damage conditions, each of which involved removing a different leg. The MAP-Elites algorithm, run for 3 million iterations, was used to create the behavior-performance maps for each of the behavioral descriptors (using a simulation of the undamaged robot). During the generation of the behavior-performance maps, the behaviors were stored in the map's cells by discretizing each dimension of the behavioral descriptor space with these five values: $\{0, 0.25, 0.5, 0.75, 1\}$ for the 6-dimensional behavioral descriptors, and with twenty equidistant values between $[0, 1]$ for the 3-dimensional behavioral descriptor. During the adaptation phase, the behaviors were used with their actual values and thus not discretized. 

We independently generated eight maps for each of the $11$ intentionally chosen behavioral descriptors. Twenty independently generated maps were generated for the random  behavioral descriptor. We launched ten replicates of each descriptor for each of the maps (eight for intentionally chosen behavioral descriptors and twenty for random behavioral descriptor) and each of the six damage conditions. There are therefore $10\times8\times6=480$ replicates for each of the intentionally chosen descriptors, and $10\times20\times6=1200$ replicates for the random descriptor. In all these simulated experiments, to roughly simulate the distribution of noisy odometry measurements on the real robot, the simulated performance values were randomly perturbed with a multiplicative Gaussian noise centered on $0.95$ with a standard deviation of $0.1$.

We analyze the fastest walking speed achieved with each behavioral descriptor after two different numbers of trials: the first case is after $17$ trials, and the second case is after $150$ trials.

\paragraph{Results}
The following results include 17 trials on the simulated robot, which was the maximum number of trials required for Intelligent Trial and Error to find a compensatory gait in the Supplementary Experiment S2. The post-adaptation performance achieved with our alternative, intentionally chosen behavioral descriptors (numbers 2-11) was similar to the original duty factor behavioral descriptor (number 1) (Extended Data Fig. \ref{fig:alt_bd}a). All $11$ alternative, intentionally chosen descriptors (numbers 2-11) led to a median performance within $17\%$ of the duty factor descriptor (performance: 0.241 [0.19; 0.29] m/s). The difference in performance was effectively nonexistent with the deviation descriptor (0.241 [0.14; 0.31] m/s), the total GRF descriptor (0.237 [0.15; 0.30] m/s), and the lower-leg roll angle descriptor (0.235 [0.14; 0.31] m/s). The lowest performance was discovered with the relative GRF descriptor ($16.7\%$ lower than the duty factor descriptor, 0.204 [0.08; 0.31] m/s). In terms of statistical significance, the performance achieved with the duty factor descriptor was no different from the deviation ($p=0.53$) and total GRF ($p=0.29$) descriptors. With all the remaining descriptors, the difference in performance was statistically significant ($p < 10^{-3}$), but it did not exceed $0.04$m/s. Additionally, the compensatory behaviors discovered with all our $11$ alternative, intentionally chosen descriptors were always faster than the reference gait for all damage conditions. 

To check whether our alternative, intentionally chosen behavioral descriptors lead to better performance if allowed a higher number of evaluations, we extended the experiments to 150 trials on the robot (Extended Data Fig. \ref{fig:alt_bd}b). After 150 trials, the difference in performance between the duty factor behavioral descriptor (0.277 [0.24; 0.34] m/s) and our alternative behavioral descriptors was further reduced. For all but three alternative, intentionally chosen descriptors (displacement, total GRF and lower-leg yaw angle), the median performance was within $4\%$ of the duty factor descriptor. The difference in performance was at $\pm3.6\%$ with the orientation (0.274 [0.22; 0.32] m/s), total energy (0.274 [0.19; 0.33] m/s), relative energy (0.273 [0.20; 0.32] m/s), deviation (0.287 [0.21; 0.34] m/s), relative GRF (0.266 [0.15; 0.35] m/s), lower-leg pitch angle (0.271 [0.21; 0.34] m/s) and lower-leg roll angle (0.268 [0.17; 0.34] m/s) descriptors. In the three remaining behavioral descriptors, displacement, total GRF, and lower-leg yaw angle, the performance was 0.264 [0.18; 0.32] m/s, 0.299 [0.25; 0.35] m/s and 0.255 [0.18; 0.32] m/s, respectively (difference at $\pm 7.8\%$ of duty factor descriptor in all three cases). In terms of statistical significance, the performance achieved with the duty factor descriptor was barely statistically significantly different from the deviation descriptor ($p=0.041$). In all the remaining descriptors, the performance difference was statistically significant ($p < 10^{-2}$), but no larger than $0.02$m/s.

Our random behavioral descriptor also performed similarly to the duty factor descriptor. After 17 trials, the performance of M-BOA with the maps generated by the random descriptor was 0.232 [0.14; 0.30] m/s ($4.2\%$ lower than the duty factor descriptor performance). While the difference is statistically significant ($p < 10^{-3}$), the difference in performance itself was negligible at $0.01$m/s. This difference in performance was further reduced to $3.6\%$ after 150 trials (random descriptor performance: 0.274 [0.21; 0.34] m/s, duty factor description performance: 0.277 [0.24; 0.34] m/s, $p=0.002$). Moreover, as with the intentionally chosen behavioral descriptors, the compensatory behavior discovered with the random descriptor was also faster than the reference gait. 

These experiments show that the selection of the behavioral dimensions is not critical to get good results. Indeed, all tested behavioral descriptors, even those randomly generated, perform well (median $>0.20$ m/s after 17 trials). On the other hand, if the robot's designers have some prior knowledge about which dimensions of variation are likely to reveal different types of behaviors, the algorithm can benefit from this knowledge to further improve results (as with the duty factor descriptor).

\section{Caption for Supplementary Videos}

\subsection*{Video S1}
\emph{This video can be viewed at: }{\footnotesize \url{https://youtu.be/T-c17RKh3uE}}\\
\textbf{Damage Recovery in Robots via Intelligent Trial and Error.}
The video shows the Intelligent Trial and Error Algorithm in action
with the two robots introduced in this paper: the hexapod robot and
the 8 degrees of freedom robotic arm (Fig. \ref{fig:results}). The
video shows several examples of the different types of behaviors that
are produced during the behavior-performance map creation step, from
classic hexapod gaits to more unexpected forms of locomotion. Then, it
shows how the hexapod robot uses that behavior-performance map to deal
with a leg that has lost power (Fig. \ref{fig:results}a:C3). Finally,
the video illustrates how the Intelligent Trial and Error Algorithm
can be applied to the second robot and to different damage conditions.

\bigskip
\subsection*{Video S2}
\emph{This video can be viewed at: }{\footnotesize \url{http://youtu.be/ycLspV5lXK8}}\\
\textbf{A Behavior-Performance Map Containing Many Different Types of
  Walking Gaits.} 
In the behavior-performance map creation step, the MAP-Elites algorithm produces a collection of different types of walking gaits. The video shows several examples of the different types of behaviors that are produced, from classic hexapod gaits to more unexpected forms of locomotion.

\balance

\section*{Supplementary References}
\begin{footnotesize}
\printbibliography[segment=2]

\end{footnotesize}
\end{refsegment}

\end{document}